\documentclass{article}

\newif\ifarxiv
\arxivtrue 
\newcommand{\onlyarxiv}[1]{\ifarxiv#1\fi}

\usepackage[T1]{fontenc}
\usepackage{microtype}
\usepackage{graphicx}
\usepackage{subcaption}
\usepackage{booktabs}
\usepackage{siunitx}
\usepackage[dvipsnames]{xcolor}
\usepackage{placeins}
\usepackage{colortbl}
\usepackage{longtable}
\usepackage{pgfplots}
\usepackage{pgfplotstable}
\usepackage{listings}

\usepackage{array}
\newcommand{\EscapeBackslash}[1]{\let\temp=\\#1\let\\=\temp}
\newcolumntype{C}[1]{>{\EscapeBackslash\centering}p{#1}}
\newcolumntype{R}[1]{>{\EscapeBackslash\raggedleft}p{#1}}
\newcolumntype{L}[1]{>{\EscapeBackslash\raggedright}p{#1}}

\usepackage{hyperref}

\usepackage[accepted]{icml2025}

\usepackage{amsmath}
\usepackage{amssymb}
\usepackage{mathtools}
\usepackage{amsthm}

\usepackage[capitalize,noabbrev]{cleveref}

\renewcommand{\vec}[1]{\mathbf{#1}}
\newcommand{\mat}[1]{\mathbf{#1}}
\newcommand{\Reals}{\mathbb{R}}

\newcommand{\W}{\mat{W}}

\newcommand{\Wenc}{\W_\text{\!enc}}
\newcommand{\Wdec}{\W_\text{\!dec}}

\newcommand{\Win}{\W_\text{\!1}}
\newcommand{\Wout}{\W_\text{\!2}}
\newcommand{\Wgate}{\W_\text{\!g}}

\newcommand{\setx}{\mathcal{X}}
\newcommand{\sety}{\mathcal{Y}}
\newcommand{\x}{\vec{x}}
\newcommand{\y}{\vec{y}}

\newcommand{\hatx}{\vec{\hat{x}}}
\newcommand{\haty}{\vec{\hat{y}}}

\newcommand{\setsx}{\mathcal{S}_\text{x}}
\newcommand{\setsy}{\mathcal{S}_\text{y}}
\newcommand{\sx}{\vec{s}_\text{x}}
\newcommand{\sy}{\vec{s}_\text{y}}

\newcommand{\sxj}{s_{\text{x},j}}
\newcommand{\syi}{s_{\text{y},i}}

\newcommand{\barsx}{\bar{\vec{s}}_\text{x}}

\newcommand{\encx}{e_\text{x}}
\newcommand{\ency}{e_\text{y}}
\newcommand{\decx}{d_\text{x}}
\newcommand{\decy}{d_\text{y}}

\newcommand{\Wencx}{\mathbf{W}_\text{x}^\text{enc}}
\newcommand{\bencx}{\mathbf{b}_\text{x}^\text{enc}}
\newcommand{\Wdecx}{\mathbf{W}_\text{x}^\text{dec}}
\newcommand{\bdecx}{\mathbf{b}_\text{x}^\text{dec}}

\newcommand{\Wency}{\mathbf{W}_\text{y}^\text{enc}}
\newcommand{\bency}{\mathbf{b}_\text{y}^\text{enc}}

\newcommand{\Wdecxmj}{W_{\text{x},mj}^\text{dec}}
\newcommand{\Wencyik}{W_{\text{y},ik}^\text{enc}}

\newcommand{\Wencyactive}{\mathbf{W}_\text{y}^\text{enc(active)}}
\newcommand{\Wdecxactive}{\mathbf{W}_\text{x}^\text{dec(active)}}

\newcommand{\dimx}{m_\text{x}}
\newcommand{\dimsx}{n_\text{x}}
\newcommand{\dimy}{m_\text{y}}
\newcommand{\dimsy}{n_\text{y}}

\newcommand{\topk}{\tau_k}

\newcommand{\fsij}{f_{s,(i,j)}|_{\sx}}

\theoremstyle{plain}

\theoremstyle{definition}

\theoremstyle{remark}

\usepackage[textsize=tiny]{todonotes}

\icmltitlerunning{Jacobian Sparse Autoencoders: Sparsify Computations, Not Just Activations}

\begin{document}

\twocolumn[
\icmltitle{Jacobian Sparse Autoencoders: Sparsify Computations, Not Just Activations}

\icmlsetsymbol{equal}{*}

\begin{icmlauthorlist}
\icmlauthor{Lucy Farnik}{bris}
\icmlauthor{Tim Lawson}{bris}
\icmlauthor{Conor Houghton}{bris}
\icmlauthor{Laurence Aitchison}{bris}
\end{icmlauthorlist}

\icmlaffiliation{bris}{School of Engineering Mathematics and Technology, University of Bristol, Bristol, UK}

\icmlcorrespondingauthor{Lucy Farnik}{lucyfarnik@gmail.com}

\icmlkeywords{Mechanistic Interpretability, Sparse Autoencoders, Superposition, Circuits, AI Safety}

\vskip 0.3in
]

\printAffiliationsAndNotice{}


\begin{abstract}
Sparse autoencoders (SAEs) have been successfully used to discover sparse and human-interpretable representations of the latent activations of LLMs.
However, we would ultimately like to understand the computations performed by LLMs and not just their representations.
The extent to which SAEs can help us understand computations is unclear because they are not designed to ``sparsify'' computations in any sense, only latent activations.
To solve this, we propose Jacobian SAEs (JSAEs), which yield not only sparsity in the input and output activations of a given model component but also sparsity in the computation (formally, the Jacobian) connecting them.
With a na\"ive implementation, the Jacobians in LLMs would be computationally intractable due to their size.
One key technical contribution is thus finding an efficient way of computing Jacobians in this setup.
We find that JSAEs extract a relatively large degree of computational sparsity while preserving downstream LLM performance approximately as well as traditional SAEs.
We also show that Jacobians are a reasonable proxy for computational sparsity because MLPs are approximately linear when rewritten in the JSAE basis.
Lastly, we show that JSAEs achieve a greater degree of computational sparsity on pre-trained LLMs than on the equivalent randomized LLM. This shows that the sparsity of the computational graph appears to be a property that LLMs learn through training, and suggests that JSAEs might be more suitable for understanding learned transformer computations than standard SAEs.
\end{abstract}

\section{Introduction}

Sparse autoencoders (SAEs) have emerged as a powerful tool for understanding the internal representations of large language models \citep{bricken_monosemanticity_2023,cunningham_sparse_2023,gao_scaling_2024,rajamanoharan_jumping_2024,lieberum_gemma_2024,lawson_residual_2024,braun_identifying_2024,kissane2024interpretingattentionlayeroutputs,rajamanoharan_improving_2024}.
By decomposing neural network activations into sparse, interpretable components, SAEs have helped researchers gain significant insights into how these models process information \citep{marks_sparse_2024,lieberum_gemma_2024,templeton_scaling_2024,obrien_steering_2024,farrell2024applyingsparseautoencodersunlearn,paulo_automatically_2024,balcells2024evolutionsaefeatureslayers,lan2024sparseautoencodersrevealuniversal,brinkmann2025largelanguagemodelsshare, spies2024transformersusecausalworld}.

When trained on the activation vectors from neural network layers, SAEs learn to reconstruct the inputs using a dictionary of sparse `features', where there are many more features than basis dimensions of the inputs, and each feature tends to capture a specific, interpretable concept.
However, the goal of this paper is to improve understanding of \emph{computations} in transformers.
While SAEs are designed to disentangle the representations of concepts in the LLM, they are not designed to help us understand the computations performed with those representations.

One approach to understanding computation would be to train two SAEs, one at the input and one at the output of an MLP in a transformer. 
We can then ask how the MLP maps sparse latent features at the inputs to sparse features in the outputs.
For this mapping to be interpretable, it would be desirable that it is sparse, in the sense that each latent in the SAE trained on the output depends on a small number of latents of the SAE trained on the input.
These dependencies can be understood as a computation graph or `circuit' \citep{olah_zoom_2020,cammarata_thread_2020}.
SAEs are not designed to encourage this computation graph to be sparse.
To address this, we develop Jacobian SAEs (JSAEs), where we include a term in the objective to encourage SAE bases with sparse computational graphs, not just sparse activations.
Specifically, we treat the mapping between the latent activations of the input and output SAEs as a function and encourage its Jacobian to be sparse by including an $L^1$ penalty term in the loss function.


\begin{figure*}
    \centering
    \includegraphics{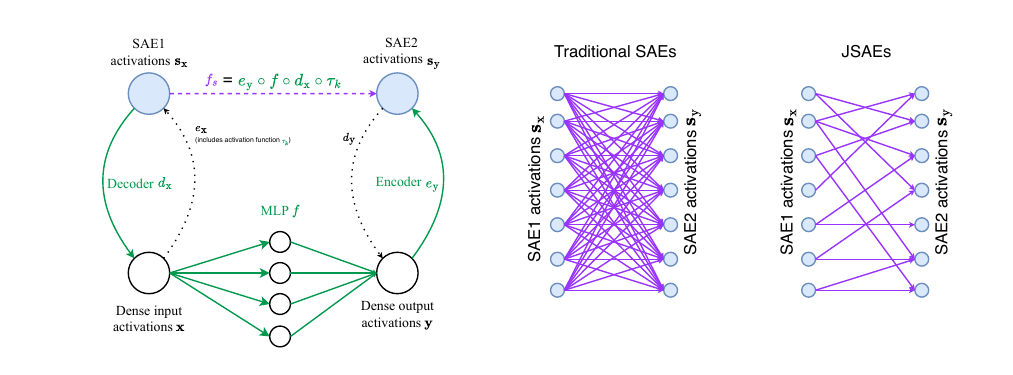}
    \caption{A diagram illustrating our setup.
    We have two SAEs: one trained on the MLP inputs and the other trained on the MLP outputs.
    We then consider the function $f_s$, which takes the latent activations of the first SAE and returns the latent activations of the second SAE, i.e., $f_s(\vec{s}_\text{x})=\vec{s}_\text{y}$.
    The function $f_s$ is described by the function composition of the TopK activation function of the first (input) SAE $\topk$, the decoder of the first SAE $\decx$, the MLP $f$, and the encoder of the second (output) SAE $\ency$.
    We note that the activation function $\topk$ is included for computational efficiency only; see Section~\ref{sec:jacobian_tractable} for details.
    JSAEs optimize for $f_s$ having a sparse Jacobian matrix, which we illustrate by reducing the number of edges in the computational graph that corresponds to $f_s$.
    Traditional SAEs have sparse SAE latents on either side of the MLP but a dense computational graph between them; JSAEs have both sparse SAE latents \textit{and} a sparse computational graph.
    Importantly, Jacobian sparsity approximates the computational graph notion, but, as we discuss in Section~\ref{sec:mostly_linear} and Appendix~\ref{app:not_local}, this approximation is highly accurate due to the fact that $f_s$ is a mostly linear function.}
    \label{fig:schematic}
\end{figure*}

%

With a na\"ive implementation, it is intractable to compute Jacobian matrices because each matrix would have on the order of a trillion elements, even for modestly sized language models and SAEs.
Therefore, one of our core contributions is to develop an efficient means to compute Jacobian matrices in this context.
The approach we develop makes it possible to train a pair of Jacobian SAEs with only approximately double the computational requirements of training a single standard SAE (Section~\ref{sec:jacobian_tractable}).
These methods enabled us to make three downstream findings.

First, we find that Jacobian SAEs successfully induce sparsity in the Jacobian matrices between input and output SAE latents relative to standard SAEs without a Jacobian term (Section~\ref{sec:jac_sparsity}).
We find that JSAEs achieve the desired increase in the sparsity of the Jacobian with only a slight decrease in reconstruction quality and model performance preservation, which remain roughly on par with standard SAEs.
We also find that the input and output latents learned by Jacobian SAEs are approximately as interpretable as standard SAEs, as quantified by auto-interpretability scores.
Importantly, we also find that the "computational units" discovered by JSAEs are often highly interpretable -- for example, JSAEs find an output latent corresponding to whether the text is in German, which is computed using several input latents corresponding to tokens frequently found in German text (Section~\ref{sec:max_act}).

Second, inspired by \citet{heap_sparse_2025}, we investigated the behavior of Jacobian SAEs when applied to random transformers, i.e., where the parameters have been reinitialized.
We find that the degree of Jacobian sparsity that can be achieved when JSAEs are applied to a pre-trained transformer is much greater than the sparsity achieved for a random transformer (Section~\ref{sec:random}).
This preliminary finding suggests that Jacobian sparsity may be a useful tool for discovering learned computational structure.

Lastly, we find that Jacobians accurately approximate computational sparsity in this context because the function we are analyzing (i.e., the combination of JSAEs and MLP) is approximately linear (Section~\ref{sec:mostly_linear}).

Our source code can be found at \href{https://github.com/lucyfarnik/jacobian-saes}{https://github.com/lucyfarnik/jacobian-saes}.

\section{Related work}

\subsection{Sparse autoencoders}

SAEs have been widely applied to `disentangle' the representations learned by transformer language models into a very large number of concepts, a.k.a. sparse latents, features, or dictionary elements
\citep{sharkey_taking_2022,cunningham_sparse_2023,bricken_monosemanticity_2023,gao_scaling_2024,rajamanoharan_jumping_2024,lieberum_gemma_2024}.
Human experiments and quantitative proxies apparently confirm that SAE latents are much more likely to correspond to human-interpretable concepts than raw language-model neurons, i.e., the basis dimensions of their activation vectors \citep{cunningham_sparse_2023,bricken_monosemanticity_2023,rajamanoharan_improving_2024}.
SAEs have been successfully applied to modifying the behavior of LLMs by using a direction discovered by an SAE to ``steer'' the model towards a certain concept \citep{makelov_sparse_2024,obrien_steering_2024,templeton_scaling_2024}.

Our work is based on SAEs but has a very different aim: standard SAEs only sparsify activations, while JSAEs also sparsify the computation graph between them (Figure~\ref{fig:schematic}).

\subsection{Transcoders}

In this paper, we focus on MLPs.
\citet{dunefsky_transcoders_2024,templeton_predicting_2024} developed \emph{transcoders}, an alternative SAE-like method to understand MLPs. 
However, JSAEs and transcoders take radically different approaches and solve radically different problems.
This is perhaps easiest to see if we look at what transcoders and JSAEs sparsify.
JSAEs are fundamentally an extension of standard SAEs: they train SAEs at the input and output of the MLP and add an extra term to the objective such that these sparse latents are also appropriate for interpreting the MLP (Figure~\ref{fig:schematic}).
In contrast, transcoders do not sparsify the inputs and outputs; they work with dense inputs and outputs.
Instead, transcoders, in essence, sparsify the MLP hidden states.
Specifically, a transcoder is an MLP that you train to match (using a mean squared error objective) the input-to-output mapping of the underlying MLP from the transformer.
The key difference between the transcoder MLP and the underlying MLP is that the transcoder MLP is much wider, and its hidden layer is trained to be sparse.

Thus, transcoders and JSAEs take fundamentally different approaches.
Each transcoder latent tells us `there is computation in the MLP related to [concept].'
By comparison, JSAEs learn a pair of SAEs (which have mostly interpretable latents) and sparse connections between them.
At a conceptual level, JSAEs tell us that `this feature in the MLP's output was computed using only these few input features'.
Ultimately, we believe that the JSAE approach, grounded in understanding how the SAE basis at one layer is mapped to the SAE basis at another layer, is potentially powerful and worth thoroughly exploring.

Importantly, it is worth emphasizing that JSAEs and transcoders are asking fundamentally different questions, as can be seen in terms of e.g., differences in what they sparsify.  
As such, it is not, to our knowledge, possible to design meaningful quantitative comparisons, at least not without extensive future work to develop very general auto-interpretability methods for evaluating methods of understanding MLP circuits.

%
%
%

\subsection{Automated circuit discovery}

In ``automated circuit discovery'', the goal is to isolate the causally relevant intermediate variables and connections between them necessary for a neural network to perform a given task \citep{olah_zoom_2020}.
In this context, a circuit is defined as a computational subgraph with an interpretable function.
The causal connections between elements are determined via activation patching, i.e., modifying or replacing the activations at a particular site of the model \citep{meng_locating_2022,zhang_best_2023,wang_interpretability_2022,hanna_how_2023}.
In some cases, researchers have identified sub-components of transformer language models with simple algorithmic roles that appear to generalize across models \citep{olsson_context_2022}.

\citet{conmy_automated_2023} proposed a means to automatically prune the connections between the sub-components of a neural network to the most relevant for a given task using activation patching.
Given a choice of task (i.e., a dataset and evaluation metric), this approach to automated circuit discovery (ACDC) returns a minimal computational subgraph needed to implement the task, e.g., previously identified `circuits' like \citet{hanna_how_2023}.
Naturally, this is computationally expensive, leading other authors to explore using linear approximations to activation patching \citep{nanda_attribution_2023,syed_attribution_2024,atpstar}.
\citet{marks_sparse_2024} later improved on this technique by using SAE latents as the nodes in the computational graph.

In a sense, these methods are supervised because they require the user to specify a task.
Naturally, it is not feasible to manually iterate over all tasks an LLM can perform, so a fully unsupervised approach is desirable.
With JSAEs, we take a step towards resolving this problem, although the architecture introduced in this paper initially only applies to a single MLP layer and not an entire model.
Additionally, to the best of our knowledge, no automated circuit discovery algorithm sparsifies the computations inside of MLPs.

There are also other approaches which focus on locating relevant computation in ML models by estimating the contribution of individual model components \citep{shah2024decomposingeditingpredictionsmodeling,balasubramanian2024decomposinginterpretingimagerepresentations}.

\section{Background}

\subsection{Sparse autoencoders}
In an SAE, we have input vectors, $\x \in \setx = \mathbb{R}^{\dimx}$. 
We want to approximate each vector $\x$ by a sparse linear combination of vectors, $\sx \in \setsx = \mathbb{R}^{\dimsx}$.
The dimension of the sparse vector, $\dimsx$, is typically much larger than the dimension of the input vectors $\dimx$ (i.e.\ the basis is overcomplete).

In the case of SAEs, we treat the vectors as inputs to an autoencoder with an encoder $\encx: \setx \rightarrow \setsx$ and a decoder $\decx: \setsx \rightarrow \setx$ defined by,
\begin{align}
  \sx &= \encx(\x) = \phi(\Wencx \x + \bencx)\\
  \hatx &= \decx(\sx) = \Wdecx \sx + \bdecx
\end{align}
Here, the parameters are the encoder weights $\Wenc \in \mathbb{R}^{\dimsx \times \dimx}$, decoder weights $\Wdec \in \mathbb{R}^{\dimx \times \dimsx}$, encoder bias $\bencx\in\mathbb{R}^{\dimsx}$, and decoder bias $\bdecx\in\mathbb{R}^{\dimx}$.
The non-linearity $\phi$ can be, for instance, ReLU.
These parameters are then optimized to minimize the difference between $\x$ and $\hatx$, typically measured in terms of the mean squared error (MSE), while imposing an $L^1$ penalty on the latent activations $\sx$ to incentivize sparsity.

\subsection{Automatic interpretability of SAE latents}

In order to compare the quality of different SAEs, it is desirable to be able to quantify how interpretable its latents are.
A popular approach to quantifying interpretability at scale is to collect the examples that maximally activate a given latent, prompt an LLM to generate an explanation of the concept the examples have in common, and then prompt an LLM to predict whether a given prompt activates the SAE latent given the generated explanation.
We can then score the accuracy of the predicted activations relative to the ground truth.
There are several variants of this approach \citep[e.g.,][]{bills_language_2023,choi_scaling_2024}; in this paper, we use ``fuzzing'' where the scoring model classifies whether the highlighted tokens in prompts activate an SAE latent given an explanation of that latent \citep{paulo_automatically_2024}.

\section{Methods}
\label{sec:methods}

The key idea with a Jacobian SAE is to train a pair of SAEs on the inputs and outputs of a neural network layer while additionally optimizing the sparsity of the Jacobian of the function that relates the input and output SAE latent activations (Figure~\ref{fig:schematic}).
In this paper, we apply Jacobian SAEs to multi-layer perceptrons (MLPs) of the kind commonly found in transformer language models \citep{radford_language_2019,biderman_pythia_2023}.

\subsection{Setup}
\label{sec:methods_setup}

Consider an MLP mapping from $\x \in \setx$ to $\y \in\sety$, i.e., $f: \setx \rightarrow \sety$ or $\y = f(\x)$.
We can then train two $k$-sparse SAEs, one on $\x$ and the other on $\y$.
The resulting SAEs map from each of $\x$ and $\y$ to corresponding sparse latents $\sx \in \setsx$ and $\sy \in \setsy$, i.e., $\sx = \encx(\x)$ and $\sy=\ency(\y)$, where $\encx$ is the encoder of the first SAE and $\ency$ is the encoder of the second SAE. 
Each of these SAEs also has a decoder that maps from the sparse latents back to an approximation of the original vector: $\hatx = \decx(\sx)$ and $\haty = \decy(\sy)$.

We may now consider the function $f_s:S_X \to S_Y$, which intuitively represents the function, $f$, but written in terms of the sparse bases learned by the SAE pair for the original vectors $\x$ and $\y$.
Specifically, we define $f_s$ by
\begin{align}
f_s=\ency \circ f\circ \decx \circ \topk
\end{align}
where $\circ$ denotes function composition.
Here, $\decx: \setsx \rightarrow \setx$ maps the sparse latents given as input to $f_s$ to ``dense'' inputs. Then, $f: \setx \rightarrow \sety$ maps the dense inputs to dense outputs. 
Finally, $\ency: \sety \rightarrow \setsy$ maps the dense outputs to sparse outputs.
Note that $f_s$ first applies the TopK activation function $\topk$ to the sparse inputs, $\sx$.
Critically, with $k$-sparse SAEs, we produce the sparse inputs by $\sx = \encx(\x)$, implying that $\sx$ only has $k$ non-zero elements. 
In that setting, TopK does not change the inputs, i.e.\ $\sx = \topk(\sx)$, but it does affect the Jacobian and, in particular, allows us to compute it much more efficiently (Section~\ref{sec:jacobian_tractable}).

At a high level, we want the function $f_s$ to be `sparse', in the sense that each of its input dimensions (i.e. SAE latent activations) only affects a small number of its output dimensions, and each of its output dimensions only depends on a small number of its input dimensions.
We quantify the sparsity of $f_s$ in terms of its Jacobian matrix.
The Jacobian of $f_s$ is, in index notation:
\begin{align}
  J_{f_s,i,j} &= \frac{\partial f_{s,i}(s_\text{x})}{\partial s_{\text{x},j}}.
\end{align}
Intuitively, we can consider maximizing the sparsity of the Jacobian as minimizing the number of edges in the computational graph connecting the input and output nodes (Figure~\ref{fig:schematic}), i.e.\ maximizing the number of near-zero elements in the Jacobian matrix.
We note that the Jacobian is not a perfect measure of the sparsity of the computational graph, but it is an accurate proxy (see Section~\ref{sec:mostly_linear} and Appendix \ref{app:not_local}) while being computationally tractable.

We simultaneously train two separate SAEs on the input and output of a transformer MLP with the objectives of low reconstruction error and sparse relations between the separate SAE latents (via the Jacobian).
We do not need to optimize for the sparsity of the latent activations via a penalty term in the loss function because we use $k$-sparse autoencoders, which keep only the $k$ largest latent activations per token position.
Hence, our loss function is
\begin{equation}
   \mathcal{L} = \text{MSE}(\x,\hatx)
   + \text{MSE}(\y,\haty) 
   + \frac{\lambda}{k^2}\sum_{i=1}^{\dimsy} \sum_{j=1}^{\dimsx} |J_{f_s,i,j}|
\end{equation}
Here, $k$ is the number of non-zero elements in the TopK activation function, $\dimsx,\dimsy$ are the dimensionalities of the latent spaces of the input and output SAEs, respectively, and $\lambda$ is the coefficient of the Jacobian loss term.
We divide by $k^2$ because, as we will see later, there are at most $k^2$ non-zero elements in the Jacobian.
Finally, note that if we set $\lambda=0$, then our objective effectively trains traditional SAEs for each of $\x$ and $\y$ independently.

\subsection{Making the Jacobian calculation tractable}
\label{sec:jacobian_tractable}

\begin{figure}[t]
    \centering
    \includegraphics{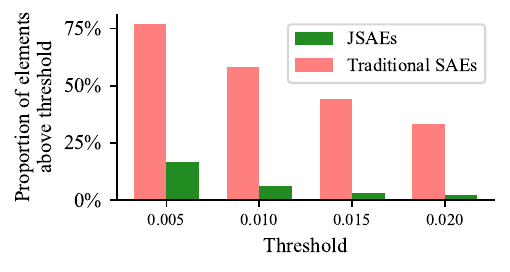}
    \caption{
    JSAEs induce a much greater degree of sparsity in the elements of the Jacobian of $f_s$ than traditional SAEs.
    The bars show the average proportion of Jacobian elements with absolute values above certain thresholds. At most $k\times k$ elements can be nonzero, so we take 100\% on the y-axis to mean $k\times k$.
    The average was taken across 10 million tokens.
    This example is from layer 15 of Pythia-410m. For layer 3 of Pythia-70m and layer 7 of Pythia-160m, see Figure~\ref{fig:jac_hist_multiple_models}, for more quantitative information on Jacobian sparsity across model sizes, layers, and hyperparameters see Figures~\ref{fig:expansion_factor}, \ref{fig:k}, and \ref{fig:jacobian_coef_70m}.
    We present further discussion of the sparsity of the Jacobian in Appendix~\ref{app:jac_sparsity}.
    }
    \label{fig:jac_sparsity}
\end{figure}

\begin{figure}[t]
    \centering
    \includegraphics{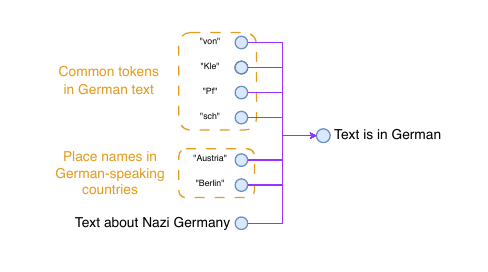}
    \caption{JSAEs allow us to locate the "input features" of each feature computed by the MLP. For instance, in Pythia-410m, the MLP at layer 15 is computing the feature "this text is in German". JSAEs discover the inputs which the MLP uses to decide whether this feature should be on or off. These inputs correspond to tokens frequently found in German text, place names in German-speaking countries, and text about Nazi Germany. See Appendix~\ref{app:qualitative} for details.}
    \label{fig:jsaes-qualitative-german}
\end{figure}

Computing the Jacobian naively (e.g., using an automatic differentiation package) is computationally intractable, as the full Jacobian has size $B\times \dimsy\times \dimsx$ where $B$ is the number of tokens in a training batch $\dimsx$ is the number of SAE latents for the input, and $\dimsy$ is the number of SAE latents for the output.
Unfortunately, typical values are around $1,000$ for $B$ and around $32,000$ for $\dimsx$ and $\dimsy$ (taking as an example a model dimension of $1,000$ and an expansion factor of 32).
Combined, this gives a Jacobian with around 1 trillion elements.
This is obviously far too large to work with in practice, and our key technical contribution is to develop an efficient approach to working with this huge Jacobian.

Our first insight is that for each element of the batch, we have a $\dimsy \times \dimsx$ Jacobian, where $\dimsx$ and $\dimsy$ are around $32,000$.
This is obviously far too large.
However, remember that we are interested in the Jacobian of $f_s$, so the input is the sparse SAE latent vector, $\sx$ and the output is the sparse SAE latent vector, $\sy$.
Importantly, as we are using $k$-sparse SAEs, only $k$ elements of the input and output are ``on'' for any given token.
As such, we really only care about the $k \times k$ elements of the Jacobian of $f_s$, corresponding to the inputs and outputs that are ``on''.
This reduces the size of the Jacobian by around six orders of magnitude, and renders the computation tractable.
However, to make this work formally, we need all elements of the Jacobian corresponding to ``off'' elements of the input and output to be zero.
This is where the $\topk$ in the definition of $f_s$ becomes important.
Specifically, the $\topk$ ensures that the gradient of $f_s$ wrt any of the inputs that are ``off'' is zero.
Without $\topk$, the Jacobian could be non-zero for any of the inputs, even if changing those inputs would not make sense, as it would give more than $k$ elements being ``on'' in the input, and thus could not be produced by the k-sparse SAE.


Our second insight was that computing the Jacobian by automatic differentiation would still be relatively inefficient, e.g., requiring $k$ backward passes.
Instead, for standard GPT-2-style MLPs, we noticed that an extremely efficient Jacobian formula can be derived by hand, requiring only three matrix multiplications and along with a few pointwise operations.
We present this derivation in Appendix~\ref{app:jac-comp}.


With these optimizations in place, training a pair of JSAEs takes about twice as long as training a single standard SAE.
We measured this by training ten of each model on Pythia-70m with an expansion factor of 32 for 100 million tokens on an RTX 3090. The average training durations were 72mins for a pair of JSAEs and 33 mins for a traditional SAE, with standard deviations below 30 seconds for both.

\section{Results}

Our experiments were performed on LLMs from the Pythia suite \citep{biderman_pythia_2023}, the figures in the main text contain results from Pythia-410m unless otherwise specified.
We trained on 300 million tokens with $k=32$ and an expansion factor of $64$ for Pythia-410m and $32$ for smaller models.
We reproduced all our experiments on multiple models and found the same qualitative results (see Appendix~\ref{app:evaluation}).

\subsection{Jacobian sparsity, reconstruction quality, and auto-interpretability scores}\label{sec:jac_sparsity}

\begin{figure*}
    \centering
    \includegraphics{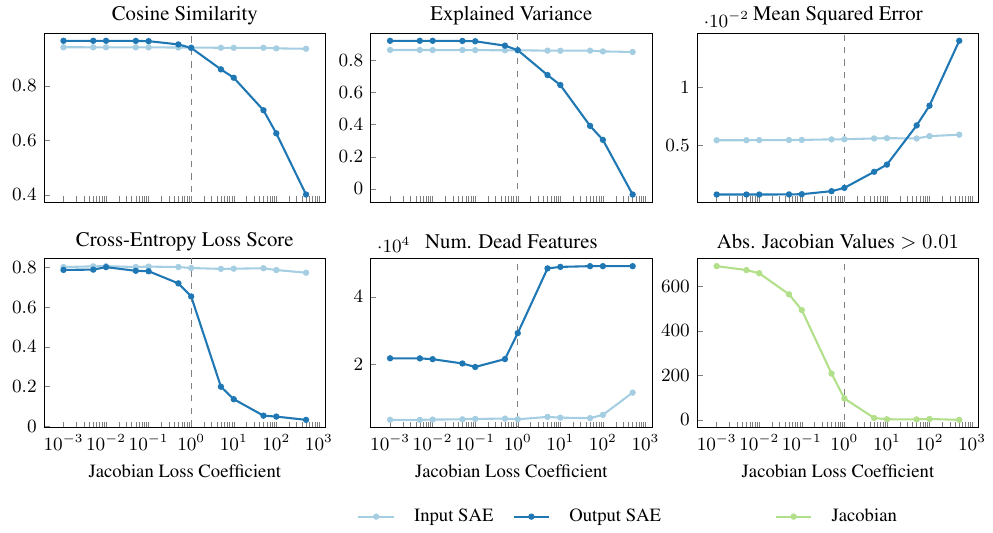}
    \caption{
        Reconstruction quality, model performance preservation, and sparsity metrics against the Jacobian loss coefficient.
        JSAEs trained on layer 7 of Pythia-160m with expansion factor $64$ and $k=32$; see Figure~\ref{fig:jacobian_coef_70m} for layer 3 of Pythia-70m.
        Recall that the maximum number of non-zero Jacobian values is $k^2=1024$.
        In accordance with Figure~\ref{fig:tradeoff_reconstr_sparsity_410m}, all evaluation metrics degrade for values of the coefficient above 1.
        See Appendix~\ref{app:evaluation} for details of the evaluation metrics.
    }
    \label{fig:jacobian_coef_160m}
\end{figure*}

First, we compared the Jacobian sparsity for standard SAEs and JSAEs.
Note that, unlike with SAE latent activations, there is no mechanism for producing exact zeros in the Jacobian elements corresponding to active latents.
Hence, we consider the number of near-zero elements rather than the number of exact zeros.
To quantify the difference in sparsity between the two, we looked at the proportion of the elements of the Jacobian above a particular threshold when aggregating over 10 million tokens (Figure~\ref{fig:jac_sparsity}).
Here, we found that JSAEs dramatically reduced the number of large elements of the Jacobian relative to traditional SAEs.
We also note that the Jacobians are not only sparse on each individual token, but also when averaged across a large number of tokens (see Figure~\ref{fig:jac_sparsity_global} in the appendix).

Importantly, the degree of sparsity depends on our choice of the coefficient $\lambda$ of the Jacobian loss term.
Therefore, we trained multiple JSAEs with different values of this parameter.
As we might expect, for small values of $\lambda$, i.e., little incentive to sparsify the Jacobian, the input and output SAEs perform similarly to standard SAEs (Figure~\ref{fig:jacobian_coef_160m} blue lines), including in terms of the variance explained by the reconstructed activation vectors and the increase in the cross-entropy loss when the input activations are replaced by their reconstructions.
Unsurprisingly, as $\lambda$ grows larger and the Jacobian loss term starts to dominate, our evaluation metrics degrade.
Interestingly, this degradation happens almost entirely in the output SAE rather than the input SAE --- we leave it to future work to investigate this phenomenon further.

\begin{figure}[t]
    \centering
    \includegraphics[width=\linewidth]{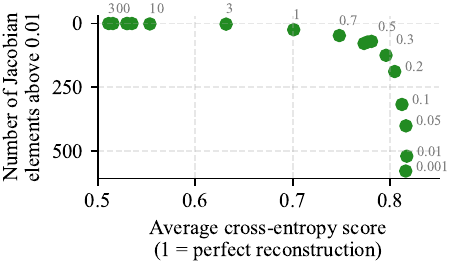}
    \caption{The trade-off between reconstruction quality and Jacobian sparsity as we vary the Jacobian loss coefficient. Each dot represents a pair of JSAEs trained with a specific Jacobian coefficient.
    The value of $\lambda$ is included for some points.
    We can see that a coefficient of roughly $\lambda=0.5$ is optimal for Pythia-410m with $k=32$.
    Note that the CE loss score is the average of the CE loss scores of the pre-MLP JSAE and the post-mlp JSAE.
    Measured on layer~15 of Pythia-410m, similar charts with a wider range of models and metrics can be found in Figures \ref{fig:pareto_70m}, \ref{fig:pareto_160m}, and \ref{fig:tradeoff_reconstr_sparsity_70m}.}
    \label{fig:tradeoff_reconstr_sparsity_410m}
\end{figure}

Critically, Figure~\ref{fig:jacobian_coef_160m} suggests there is a `sweet spot' of the $\lambda$ hyperparameter where the SAE quality metrics remain reasonable, but the Jacobian is much sparser than for standard SAEs.
To further investigate this trade-off, we plotted a measure of Jacobian sparsity (the proportion of elements of the Jacobian above 0.01) against the average cross-entropy (Figures ~\ref{fig:jacobian_coef_160m}, \ref{fig:tradeoff_reconstr_sparsity_410m}, and \ref{fig:tradeoff_reconstr_sparsity_70m}).
We found that there is indeed a sweet spot where the average cross-entropy is only slightly worse than a traditional SAE, while the Jacobian is far sparser.
For Pythia 410m (Figure~\ref{fig:tradeoff_reconstr_sparsity_410m}) this value is around $\lambda=0.5$, whereas for Pythia-70m, it is around $\lambda=1$ (Figure~\ref{fig:tradeoff_reconstr_sparsity_70m}).
We choose this value of the Jacobian coefficient (i.e.\ $\lambda=0.5$ for Pythia-410m in the main text, and $\lambda=1$ for Pythia-160m in the Appendix) in other experiments.

\begin{figure}[t]
    \centering
    \includegraphics{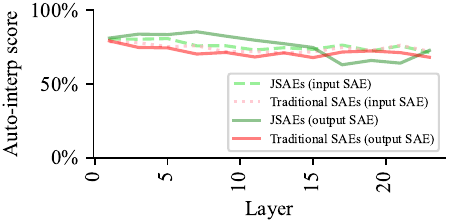}
    \caption{Automatic interpretability scores of JSAEs are very similar to traditional SAEs. Measured on all odd-numbered layers of Pythia-410m using the ``fuzzing'' scorer from \citet{paulo_automatically_2024}. For all layers of Pythia-70m see Figure~\ref{fig:autointerp_70m}.}
    \label{fig:autointerp_410m}
\end{figure}

We also measure the interpretability of JSAE latents using the automatic interpretability pipeline developed by \citet{paulo_automatically_2024} and compare this to traditional SAEs.
We find that JSAEs achieve similar interpretability scores
(Figure~\ref{fig:autointerp_410m}).

\begin{figure}[t]
    \centering
    \includegraphics{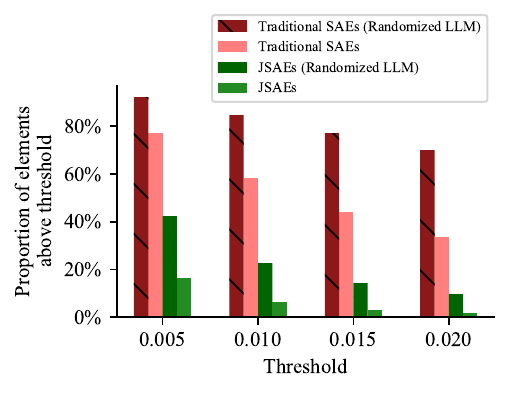}
    \caption{
    Jacobians are substantially more sparse in pre-trained LLMs than in randomly initialzied transformers.
    This holds both when you actively optimize for Jacobian sparsity with JSAEs, and when you don't optimize for it and use traditional SAEs.
    The figure shows the proportion of Jacobian elements with absolute values above certain thresholds. At most $k^2$ elements can be nonzero, we therefore take $k^2$ to be 100\% on the y-axis.
    Jacobians are significantly more sparse in pre-trained transformers than in randomly re-initialized transformers.
    This shows that Jacobian sparsity is, at least to some extent, connected to the structures that LLMs learn during training.
    This stands in contrast to recent work by \citet{heap_sparse_2025} showing that traditional SAEs achieve roughly equal auto-interpretability scores on randomly initialized transformers as they do on pre-trained LLMs.
    Measured on layer 15 of Pythia-410m, for layer 3 of Pythia-70m see Figure~\ref{fig:randomized_70m}.
    Averaged across 10 million tokens.}
    \label{fig:randomized_410m}
\end{figure}

\subsection{Max-activating examples of JSAEs}\label{sec:max_act}
Next, we interpreted the "max-activating" examples of JSAEs in order to verify that JSAEs can locate semantically meaningful computational units.
Namely, we took the latents of the output SAE $i$ which have large Jacobian values when averaging across a wide distribution of text.
Then for each output SAE latent $i$, we found the 10 input SAE latents $j$ which have the largest average Jacobian elements $J_{f_s,i,j}$.
We find that these combinations are often highly interpretable.
For example, as shown in Figure~\ref{fig:jsaes-qualitative-german}, the very first output latent of layer 15 of Pythia-410m as sorted by average Jacobian value corresponds to "this text is in German".
We find that it is computed as a function of input latents corresponding to:
\begin{itemize}
    \item Tokens which frequently appear in German text, such as "Pf", "sch", "Kle", and "von"
    \item Names of places where people speak German, such as "Berlin" or "Austria"
    \item Words and phrases related to the Third Reich, such as "Nazi", "concentration camp", "Hitler", and "Holocaust"
\end{itemize}
For a few of handpicked examples, see Appendix~\ref{app:qualitative}.
A large number of examples which are not handpicked is available at \href{https://tinyurl.com/jsaes-qualitative}{tinyurl.com/jsaes-qualitative}.




\begin{figure*}[t]
    \centering
    \includegraphics{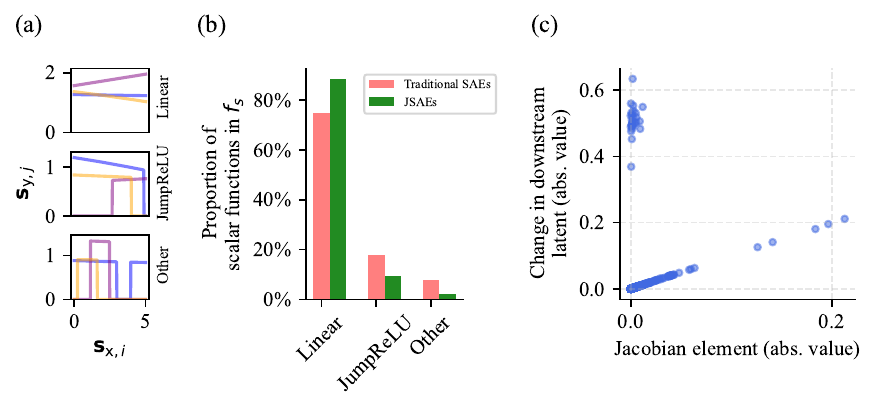}
    \caption{The function $f_s$, which combines the decoder of the first SAE, the MLP, and the encoder of the second SAE, is mostly linear.
    Specifically, the vast majority of scalar functions going from $\sxj$ to $\syi$ are linear.
    (a)~Examples of linear, JumpReLU, and other functions relating individual input SAE latents and output SAE latents. See Figure~\ref{fig:func_examples} for more examples.
    (b)~For the empirically observed $\sx$ and randomly selected $i,j$ (of those corresponding to active SAE latents), the vast majority of scalar functions from $\sxj$ to $\syi$ are linear. For details see Appendix~\ref{app:not_local}.
    The proportion of linear function also noticeably increases with JSAEs compared to traditional SAEs, meaning that JSAEs induce additional linearity in $f_s$.
    (c)~Because the vast majority of functions are linear, the Jacobian usually precisely predicts the change observed in the output SAE latent when we make a large change to the input SAE latent's value (namely subtracting 1, note that the empirical median value of $\sxj$ is $2.5$).
    Each dot corresponds to an $(\sxj,\syi)$ pair.
    For 97.7\% of pairs (across a sample size of 10 million) their Jacobian value nearly exactly predicts the change we see in the output SAE latent when making large changes to the input SAE latent's activation, i.e. $|\Delta \syi|\approx|J_{{f_s},ij}|$.
    The scatter plot shows a randomly selected subset of 1,000 $(\sxj,\syi)$ pairs. 
    For further details see Appendix~\ref{app:not_local}.
    Measured on layer 15 of Pythia-410m, for layer 3 of Pythia-70m see Figure~\ref{fig:mostly_linear_70m}, for the linearity results on other models and hyperparameters see Figures~\ref{fig:linear_layer}, \ref{fig:linear_jacobian_coef}, and \ref{fig:linear_expansion_factor_k}.
    }
    \label{fig:mostly_linear_410m}
\end{figure*}

\subsection{Performance on re-initialized transformers}\label{sec:random}~
To confirm that JSAEs are extracting information about the complex learned computation, we considered a form of control analysis inspired by \citet{heap_sparse_2025}.
Specifically, we would expect that trained transformers have carefully learned specific, structured computations while randomly initialized transformers do not.
Thus, a possible desideratum for tools in mechanistic interpretability is that they ought to work substantially better when analyzing the complex computations in trained LLMs than when applied to LLMs with randomly re-initialized weights.
This is precisely what we find.
Specifically, we find that the Jacobians for trained networks are always substantially sparser than the corresponding random trained network, and this holds for both traditional SAEs and JSAEs (Figure~\ref{fig:randomized_410m}).
Further, the relative improvement in sparsity from the traditional SAE to the JSAE is much larger for trained than random LLMs, again indicating that JSAEs are extracting structure that only exists in the trained network.
Note that we also see that for traditional SAEs, there is a somewhat more sparse Jacobian for the trained than randomly initialized transformer.
This makes sense: we would hope that the traditional SAE basis is somewhat more aligned with the computation (as expressed by a sparse Jacobian) than we would expect by chance.
However, it turns out that without a ``helping hand'' from the Jacobian sparsity term, the alignment in a traditional SAE is relatively small.
Thus, Jacobian sparsity is a property related to the complex computations LLMs learn during training, which should make it substantially useful for discovering the learned structures of LLMs.

\subsection{$f_s$ is mostly linear}\label{sec:mostly_linear}

Importantly, the Jacobian is a local measure.
Thus, strictly speaking, a near-zero element of the Jacobian matrix implies only that a small change to the input SAE latent does not affect the corresponding output SAE latent.
It may, however, still be the case that a large change to the input SAE latent would change the output SAE latent.
We investigated this question and found that $f_s$ is usually approximately linear in a wide range and is often close to linear.
Specifically, of the scalar functions relating individual input SAE latents $\sxj$ to individual output SAE latents $\syi$, the vast majority are linear (Figure~\ref{fig:mostly_linear_410m}b).
This is important because, for any linear function, its local slope is completely predictive of its global shape, and therefore, a near-zero Jacobian element implies a near-zero causal relationship.
For the scalar functions which are not linear, we frequently observed they have a JumpReLU structure\footnote{By JumpReLU, we mean any function of the form $f(x)=a\text{JumpReLU}(bx+c)$. Recall that $\text{JumpReLU}(x)=x$ if $x>d$ and $0$ otherwise. $a,b,c,d \in\Reals$ are constants.} \citep{erichson2019jumpreluretrofitdefensestrategy}.
Notably, a JumpReLU is linear in a subset of its input space, so even for these scalar functions the first derivative is still an accurate measure within some range of $\sxj$ values.
It is also worth noting that with JSAEs, the proportion of linear functions is noticeably higher than with traditional SAEs, so at least to a certain extent, JSAEs induce additional linearity in the MLP.
To confirm these results, we plotted the Jacobian against the change of output SAE latent $\syi$ as we change the input SAE latent $\sxj$ by subtracting $1$ (Figure~\ref{fig:mostly_linear_410m}c)\footnote{For reference, the median value of $\sxj$ without any interventions is $2.5$.}.
We found that 97.7\% of the time, $|\Delta \syi|\approx|J_{{f_s},ij}|$.
For details see Appendix~\ref{app:not_local}.
While these results are strongly suggestive, we would caution that it is difficult to interpret them definitively as we are not evaluating the reconstruction error for a linear model fitted to the input-output relationship for the MLP latents.

\onlyarxiv{
\section{Discussion}
\label{sec:discussion}
We believe JSAEs are a promising approach for discovering computational sparsity and understanding the reasoning of LLMs.
We would also argue that an approach like the one we introduced is in some sense necessary if we want to `reverse-engineer' or `decompile' LLMs into readable source code.
It is not enough that our variables (e.g., SAE features) are interpretable; they must also be connected in a relatively sparse way.
To illustrate this point, imagine a Python function that takes as input 5 arguments and returns a single variable, and compare this to a Python function that takes 32,000 arguments.
Naturally, the latter would be nearly impossible to reason about.
Discovering computational sparsity thus appears to be a prerequisite for solving interpretability.
It is also important that the mechanisms for discovering computational sparsity be fully unsupervised rather than requiring the user to manually specify the task being analyzed.
There are existing methods for taking a specific task and finding the circuit responsible for implementing it, but these require the user to specify the task first (e.g. as a small dataset of task-relevant prompts and a metric of success).
They are thus `supervised' in the sense that they need a clear direction from the user.
Naturally, it is not feasible to manually iterate over all tasks an LLM may be performing, so a fully unsupervised approach is needed.
JSAEs are the first step in this direction.

Naturally, JSAEs in their current form still have important limitations.
They currently only work on MLPs, and for now, they only operate on a single layer at a time rather than discovering circuits throughout the entire model.
Our initial implementation also works on GPT-2-style MLPs, while most LLMs from the last few years tend to use GLUs \cite{dauphin2017languagemodelinggatedconvolutional,shazeer2020gluvariantsimprovetransformer}, though we expect it to be fairly easy to extend our setup to GLUs.
Additionally, our current implementation relies on the TopK activation function for efficient batching; TopK SAEs can sometimes encourage high-density features, so it may be desirable to generalize our implementation to work with other activation functions.
These are, however, problems that can be addressed relatively straightforwardly in future work, and we would welcome correspondence from researchers interested in addressing them.

A pessimist may argue that partial derivatives (and, therefore, Jacobians) are merely local measures.
A small partial derivative tells you that if you slightly tweak the input latent's activation, you will see no change to the output latent's activation, but it may well be the case that a large change to the input latent's activation will lead to a large change in the output latent.
Thankfully, at least in MLPs, this is not quite the case.
As we show in Section \ref{sec:mostly_linear}, $f_s$ is approximately linear, and the size of the elements of the Jacobian nearly perfectly predicts the change you see in the output latent when you make a large change to the input latent.
For a linear function, a first-order derivative at any point is perfectly predictive of the relationship between the input and the output, and thus, at least for the fraction of $f_s$ that is linear, Jacobians perfectly measure the computational relationship between input and output variables.
We further discuss this in Appendix \ref{app:not_local}.
Additionally, as we showed in Section \ref{sec:random}, Jacobian sparsity is much more present in trained LLMs than in randomly initialized ones, which indicates that it does correspond in some way to structures that were learned during training.
At a high level, a sparse computational graph necessarily implies a sparse Jacobian, but a sparse Jacobian does not in and of itself imply a sparse computational graph.
But all of these results make it seem likely that Jacobian sparsity is a good approximation of computational sparsity, and when combined with the fact that we have now developed efficient ways of computing them at scale, this leads us to believe that JSAEs are a highly useful approach.
We would, however, still invite future work to further investigate the degree to which Jacobians, and by extension JSAEs, capture the structure we care about when analyzing LLMs.

}

\section{Conclusion}

We introduced Jacobian sparse autoencoders (JSAEs), a new approach for discovering sparse computation in LLMs in a fully unsupervised way.
We found that JSAEs induce sparsity in the Jacobian matrix of the function that represents an MLP layer in the sparse basis found by JSAEs, with minimal degradation in the reconstruction quality and downstream performance of the underlying model and no degradation in the interpretability of latents.
We demonstrated that the computation found by JSAEs is often highly interpretable, allowing us to see not only the concepts computed by MLPs, but also the "input concepts" which are used to compute each "output concept".
We also found that Jacobian sparsity is substantially greater in pre-trained LLMs than in randomly initialized ones suggesting that Jacobian sparsity is indeed a proxy for learned computational structure.
Lastly, we found that Jacobians are a highly accurate measure of computational sparsity due to the fact that the MLP in the JSAE basis consists mostly of linear functions relating input to output JSAE latents.


\section*{Acknowledgements}
The authors wish to thank Callum McDougall and Euan Ong for helpful discussions.
We also thank the contributors to the open-source mechanistic interpretability tooling ecosystem, in particular the authors of SAELens \citep{bloom_jbloomaus_2023}, which formed the backbone of our codebase.
The authors wish to acknowledge and thank the financial support of the UK Research and Innovation (UKRI) [Grant ref EP/S022937/1] and the University of Bristol.
This work was carried out using the computational facilities of the Advanced Computing Research Centre, University of Bristol - http://www.bristol.ac.uk/acrc/.
We would like to thank Dr.~Stewart for funding for GPU resources.


\section*{Impact Statement}
The work presented in this paper advances the field of mechanistic interpretability.
Our hope is that interpretability will prove beneficial in making LLMs safer and more robust in ways ranging from better detection of model misuse to editing LLMs to remove dangerous capabilities.



\section*{Author contribution statement}
Conceptualization was done by LF and LA.
Derivation of an efficient way to compute the Jacobian was done by LF and LA.
Implementation of the training codebase was done by LF.
The experiments in Jacobian sparsity, auto-interpretability, reconstruction quality, and approximate linearity of $f_s$ were done by LF.
Qualitative examples of the computations found by JSAEs were done by TL.
LA and CH provided supervision and guidance throughout the project.
The text was written by LF, LA, TL, and CH.
Figures were created by LF and TL with advice from LA and CH.


\bibliography{refs}
\bibliographystyle{icml2025}

\newpage
\appendix
\onecolumn

\section{Efficiently computing the Jacobian}\label{app:jac-comp}

A simple form for the Jacobian of the function $f_s = \ency \circ f \circ \decx \circ \topk$, which describes the action of an MLP layer $f$ in the sparse input and output bases, follows from applying the chain rule.
Note that here, the subscripts $f_s$, $e_\text{y}$, etc. denote the function in question rather than vector or matrix indices.
For the GPT-2-style MLPs that we study, the components of $f_s$ are:

\begin{enumerate}
    \item \textbf{TopK}. This function takes sparse latents $\sx$ and outputs sparse latents $\barsx$. Importantly, $\sx=\barsx$. This step makes the backward pass of the Jacobian computation more efficient but does not affect the forward pass.
    \begin{align}
        \barsx &= \topk(\sx)
    \end{align}

    \item \textbf{Input SAE Decoder}. This function takes sparse latents $\barsx$ and outputs dense MLP inputs $\hatx$:
    \begin{align}
        \hatx &= \decx(\barsx) = \Wdecx \barsx + \bdecx
    \end{align}
    
    \item \textbf{MLP}. This function takes dense inputs $\hatx$ and outputs dense outputs $\y$:
    \begin{align}
        \vec{z} = \Win \hatx + \vec{b}_\text{1}
        \ ,\quad
        \y = \Wout \phi_{\text{MLP}}(\vec{z}) + \vec{b}_\text{2}
    \end{align}
    where $\phi_{\text{MLP}}$ is the activation function of the MLP (e.g., GeLU in the case of Pythia models).
    
    \item \textbf{Output SAE Encoder}. This function takes dense outputs $\y$ and outputs sparse latents $\sy$:
    \begin{align}
        \sy &= \ency(\y) = \topk\left( \Wency \y + \bency \right)
    \end{align}
\end{enumerate}

The Jacobian $\mat{J}_{f_s} \in \mathbb{R}^{\dimsy \times \dimsx}$ for a single input activation vector has the following elements, in index notation:
\begin{align}
    \label{eq:jacobian}
    J_{f_s,ij} &=
    \frac{\partial s_{\text{y},i}}{\partial s_{\text{x},j}} =
    \sum_{k \ell m n}
    \frac{\partial s_{\text{y},i}}{\partial y_k}
    \frac{\partial y_k}{\partial z_\ell}
    \frac{\partial z_\ell}{\partial \hat{x}_m}
    \frac{\partial \hat{x}_m}{\partial \bar{s}_{\text{x},n}}
    \frac{\partial \bar{s}_{\text{x},n}}{\partial s_{\text{x},j}}
\end{align}
We compute each term like so:
\begin{enumerate}
    \item \textbf{Output SAE Encoder derivative}:
    \begin{align}
        \frac{\partial s_{\text{y},i}}{\partial y_k} &=
        \tau_k'\left(
            \sum_j W_{ij}^\text{enc} y_j + b_{\text{enc},i}
        \right) \Wencyik =
        \begin{cases}
            \Wencyik & \text{if } i\in\mathcal{K}_\text{2} \\
            0        & \text{otherwise}
        \end{cases}
    \end{align}
    where $\mathcal{K}_\text{2}$ is the set of indices selected by the TopK activation function $\tau_k$ of the second (output) SAE.
    Importantly, the subscript $k$ \emph{does not} indicate the $k$-th element of $\tau_k$, whereas it \emph{does} indicate the $k$-th column of $\Wencyik$.
    
    \item \textbf{MLP derivatives}:
    \begin{align}
        \frac{\partial y_k}{\partial z_\ell} =
        W_{\text{2},k\ell} \, \phi_{\text{MLP}}'(z_\ell)
        \ ,\quad
        \frac{\partial z_\ell}{\partial \hat{x}_m} =
        W_{\text{1},\ell m}
    \end{align}
    
    \item \textbf{Input SAE Decoder derivative}:
    \begin{align}
        \frac{\partial \hat{x}_m}{\partial \bar{s}_{\text{x},n}} =
        W_{\text{x},mn}^\text{dec}
    \end{align}

    \item \textbf{TopK derivative}:
    \begin{align}
        \frac{\partial \bar{s}_{\text{x},n}}{\partial \sxj} &= \begin{cases}
            1&\text{if } j\in\mathcal{K}_\text{1}\\
            0&\text{otherwise}
        \end{cases}
    \end{align}
    where $\mathcal{K}_\text{1}$ is the set of indices (corresponding to SAE latents) that were selected by the TopK activation function $\tau_k$ of the first (input) SAE, which we explicitly included in the definition of $f_s$ above.
\end{enumerate}

When we combine all the terms:
\begin{align}
    J_{f_s,ij} &=
    \begin{cases}
        \sum_{k\ell m}
        \Wencyik \,
        W_{\text{2},k\ell} \,
        \phi_{\text{MLP}}'(z_\ell) \,
        W_{\text{1},\ell m} \,
        \Wdecxmj
        & \text{if } i\in\mathcal{K}_\text{2}\land j \in \mathcal{K}_\text{1} \\
        0 & \text{otherwise}
    \end{cases}
\end{align}

Let $\Wencyactive \in \mathbb{R}^{k \times \dimy}$ and $\Wdecxactive \in \mathbb{R}^{\dimx \times k}$ contain the active rows and columns, i.e., the rows and columns corresponding to the $\mathcal{K}_\text{2}$ or $\mathcal{K}_\text{1}$ indices respectively. The Jacobian then simplifies to:

\begin{align}
    \mat{J}_{f_s}^\text{(active)} &= \underbrace{\Wencyactive\W_{\text{2}}}_{\mathbb{R}^{k \times d_{\text{MLP}}}} \cdot 
    \underbrace{\phi_{\text{MLP}}'(\vec{z})}_{\mathbb{R}^{d_{\text{MLP}} \times d_{\text{MLP}}}} \cdot
    \underbrace{\W_\text{1} \Wdecxactive}_{\mathbb{R}^{d_{\text{MLP}} \times k}}
\end{align}

where $d_{\text{MLP}}$ is the hidden size of the MLP.
Note that $\mat{J}_{f_s}^\text{(active)}$ is of size $k\times k$, while the full Jacobian matrix $\mat{J}_{f_s}$ is of size $\dimsy\times\dimsx$.
However, $\mat{J}_{f_s}^\text{(active)}$ contains all the nonzero elements of $\mat{J}_{f_s}$, so it is all we need to compute the loss function to train Jacobian SAEs (Section~\ref{sec:methods_setup}).

\subsection{JSAEs with GLUs}
The equations above can be easily adapted to work with gated linear units (GLUs), which are significantly more common in modern LLMs than GPT-2-style MLPs.

To do this, we modify the MLP equations like so:
\begin{align}
    \vec{g} &= \Wgate \hatx + \vec{b}_\text{g}\\
    \vec{s} &= \phi_{\text{MLP}}(\vec{g})\\
    \vec{h} &= \Win \hatx + \vec{b}_\text{1}\\
    \vec{z} &= \vec{h} \odot \vec{s}\\
    \y &= \Wout \vec{z} + \vec{b}_\text{2}
\end{align}
where $\odot$ is elementwise multiplication.

We then modify the derivatives accordingly:
\begin{align}
    \frac{\partial y_k}{\partial z_\ell} &= W_{\text{2},k\ell}\\
    \frac{\partial z_\ell}{\partial \hat{x}_m} &= h_\ell \frac{\partial s_\ell}{\partial \hat{x}_m} + s_\ell \frac{\partial h_\ell}{\partial \hat{x}_m}\\
    \frac{\partial h_\ell}{\partial \hat{x}_m} &= W_{\text{1},\ell m}\\
    \frac{\partial s_\ell}{\partial g_\ell} &= \phi_{\text{MLP}}'(g_\ell)\\
    \frac{\partial g_\ell}{\partial \hat{x}_m} &= W_{\text{g},\ell m}\\
\end{align}

Combining the terms again:
\begin{align}
    J_{f_s,ij} &=
    \begin{cases}
        \sum_{k\ell m}
        \Wencyik \,
        W_{\text{2},k\ell} \,
        (h_\ell\,\phi_{\text{MLP}}'(g_\ell)\,W_{\text{g},\ell m} + s_\ell\,W_{\text{1},\ell m})\,
        \Wdecxmj
        & \text{if } i\in\mathcal{K}_\text{2}\land j \in \mathcal{K}_\text{1} \\
        0 & \text{otherwise}
    \end{cases}
\end{align}

The Jacobian is then:
\begin{align}
    \mat{J}_{f_s}^\text{(active)} &= \underbrace{\Wencyactive\W_{\text{2}}}_{\mathbb{R}^{k \times d_{\text{MLP}}}} \cdot 
    \underbrace{\left(\text{diag}\left(\vec{h}\odot\phi_{\text{MLP}}'(\vec{g})\right)\Wgate + \text{diag}(\vec{s})\mat{W}_{\text{1}}\right)}_{\mathbb{R}^{d_{\text{MLP}}} \times \dimx} \cdot
    \underbrace{\Wdecxactive}_{\mathbb{R}^{\dimx \times k}}
\end{align}

\section{$f_s$ is approximately linear}\label{app:not_local}

Consider the scalar function $\fsij:\Reals\to\Reals$ which takes as input the $j$-th latent activation of the first SAE (i.e. $\sxj$) and returns as output the $i$-th latent activation of the second SAE (i.e., $\syi$), while keeping the other elements of the input vector fixed at the same values as $\sx$.
In other words, this function captures the relationship between the $j$-th input SAE latent and the $i$-th output SAE latent in the context of $\sx$.
Geometrically, we start off at the point $\sx$, and we move from it through the input spaces in parallel to the $j$-th basis vector, and then we observe how the output of $f_s$ projects onto the $i$-th basis vector.
Formally,
\begin{align}
   \fsij(x)&=f_s\left(\psi(\sx,i,x)\right)_j \label{eq:def_fsij}\\
   \psi(\sx,i,x)_k &=\begin{cases}
       x&\text{if }i=k\\
       s_{\text{x},j} &\text{otherwise}
   \end{cases}
\end{align}
These are the functions shown in Figure~\ref{fig:mostly_linear_410m}a, of which the vast majority are linear (Figure~\ref{fig:mostly_linear_410m}b).

As we showed in Figure \ref{fig:mostly_linear_410m}c, the absolute value of a Jacobian element nearly perfectly predicts the change we see in the output SAE latent activation value when we make a large intervention on the input SAE latent activation.
However, in the same figure, there is a small cluster of approximately 2.5\% of samples, where the Jacobian element is near zero, but the change observed in the downstream feature is quite large.
We proceed by exploring the cause behind this phenomenon.

\begin{figure}
    \centering
    \includegraphics{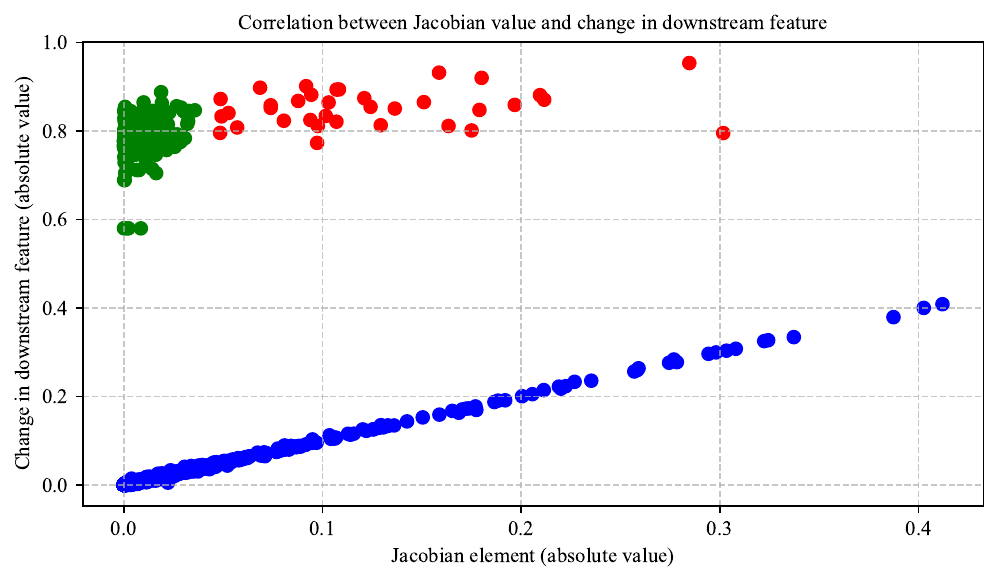}
    \caption{An expanded version of Figure~\ref{fig:mostly_linear_410m}c, measured on layer 3 of Pythia-70m. A scatter plot showing that values of Jacobian elements tend to be approximately equal to the change we see in the downstream feature when we modify the value of the upstream feature, namely when we subtract 1 from it.
    Each dot corresponds to an (input SAE latent, output SAE latent) pair.
    Unlike Figure~\ref{fig:mostly_linear_410m}c, this figure colors in the dots depending on which cluster they belong to -- blue for "on the line", green for "in the cluster", red for "outlier".
    Additionally, this figure contains 10,000 samples (rather than 1,000 as in Figure~\ref{fig:mostly_linear_410m}c), which allows us to see more of the outliers and edge cases, though at the cost of visually obfuscating the fact that 97.5\% of the samples are on the diagonal line, 2.1\% are in the cluster, and 0.4\% are outliers.
    }
    \label{fig:not_local_expanded}
\end{figure}

\begin{figure}
    \centering
    \includegraphics{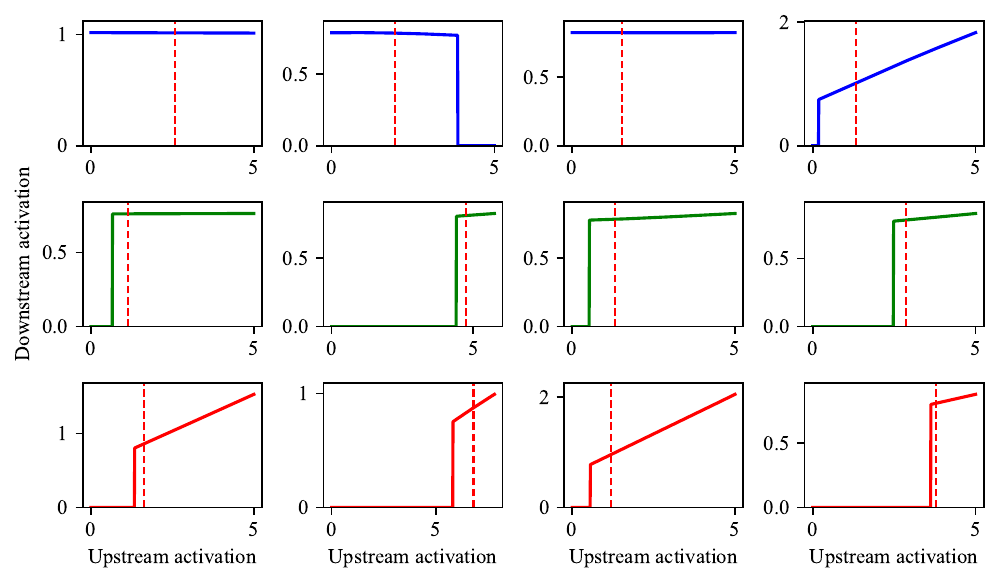}
    \caption{A handful of $\fsij$ functions corresponding to the points in Figure \ref{fig:not_local_expanded}.
    The color matches the group (and therefore the color) they were assigned in Figure \ref{fig:not_local_expanded}.
    The red dashed vertical line denotes $\vec{s}_{\text{x},i}^{(l)}$, i.e. the activation value of the SAE latent before we intervened on it.
    Note that the functions are not selected randomly but rather hand-selected to demonstrate the range of functions.
    We will quantitatively explore what proportion of $\fsij$ functions have which structure in other figures.}
    \label{fig:not_local_functions}
\end{figure}

\begin{figure}
    \centering
    \includegraphics{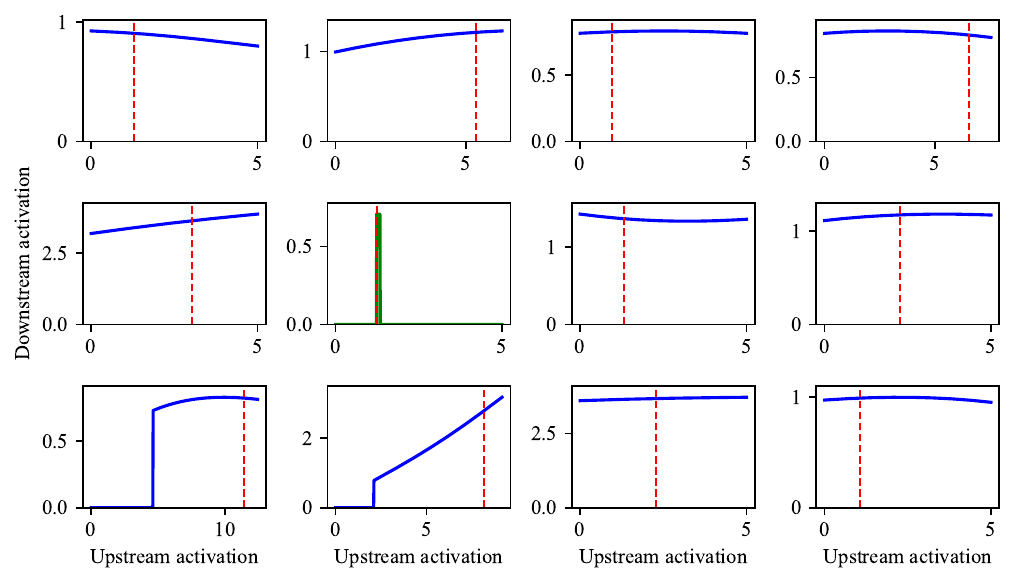}
    \caption{A random selection of the non-linear, non-JumpReLU $\fsij$ functions.
    Note that non-linear, non-JumpReLU functions only constitute about 2\% of $\fsij$ functions.
    Even though these functions are clearly somewhat non-linear, their slope does still change quite slowly for the most part, which means that a first-order derivative at any point in the function is still reasonably predictive of the function's behavior in at least some portion of the input space (though there are some rare exceptions).
    The color again matches the group (and therefore the color) they were assigned in Figure \ref{fig:not_local_expanded};
    the red dashed vertical line denotes $\vec{s}_{\text{x},i}^{(l)}$, i.e. the activation value of the SAE latent before we intervened on it.}
    \label{fig:nonlinear_functions}
\end{figure}

Note that each point in Figure~\ref{fig:mostly_linear_410m} corresponds to a single scalar function $\fsij$ (a pair of latent indices).
An expanded version of Figure \ref{fig:mostly_linear_410m} is presented in Figure \ref{fig:not_local_expanded}.
Importantly, we show the `line', the top-left cluster, and outliers visible in Figure~\ref{fig:mostly_linear_410m} in different colors, which we re-use in the following charts (Figures~\ref{fig:not_local_functions} and \ref{fig:nonlinear_functions}).
It also includes 10K samples, compared to 1K in Figure \ref{fig:mostly_linear_410m}c: as above, most samples remain on the line, but the greater number of samples makes the behavior of the top-left cluster and outliers clearer.

Figure \ref{fig:not_local_functions} illustrates some examples of functions $\fsij$ taken from each category shown in Figure~\ref{fig:mostly_linear_410m}, i.e., the line, cluster, and outliers.
The vast majority of functions belong to the line category and are typically either linear or akin to JumpReLU activation functions (which include step functions as a special case).
By contrast, the minority of functions belonging to the cluster or outliers are typically also JumpReLU-like, except where the unmodified input latent activation is close to the point where the function `jumps', so when we subtract an activation value of 1 from the input (as in Figures~\ref{fig:mostly_linear_410m}c and \ref{fig:not_local_expanded}), this moves to the flat region where the output latent activation value is zero.

As we can see, the vast majority of these functions are either linear or JumpReLUs.
\samepage{Indeed, we verify this across the sample size of 10,000 functions and find that 88\% are linear, 10\% are JumpReLU (excl. linear, which is arguably a special case of JumpReLU), and only 2\% are neither\footnote{Note that we are testing whether functions are linear or JumpReLUs only in the region of input space within which SAE activations exist. In particular, this means that we are excluding negative numbers. More specifically, the domain within which we test the function's structure is $[0,\max(5, \vec{s}_{\text{x},i}^{(l)}+1)]$. In 92\% of cases, $\vec{s}_{\text{x},i}^{(l)}+1 < 5$; the median $\vec{s}_{\text{x},i}^{(l)}$ is 2.5.}. \label{footnote:fsij_domain}}
This result is encouraging -- for a linear function, the first-order derivative is constant, so its value (i.e., the corresponding element of the Jacobian) completely expresses the relationship between the input and output values (up to a constant intercept).
For the 88\% of these scalar functions that are linear, the Jacobian thus accurately captures the notion of computational sparsity that interests us, rather than serving only as a proxy.
And for the 10\% of JumpReLUs, the Jacobians still perfectly measure the computational change we observe when changing the input latent within some subset of the input space.

While we expect the remaining 2\% of scalar functions (Jacobian elements) to contribute only a small fraction of the computational structure of the underlying model, we preliminarily investigated their behavior.
Figure \ref{fig:nonlinear_functions} shows 12 randomly selected non-linear, non-JumpReLU $\fsij$ functions.
Even though these functions are nonlinear, they are still reasonably close to being linear, i.e., their first derivative is still predictive of the change we see throughout the input space.
Indeed, most of them are on the diagonal line in Figure \ref{fig:not_local_expanded}.

\begin{figure}
    \centering
    \includegraphics{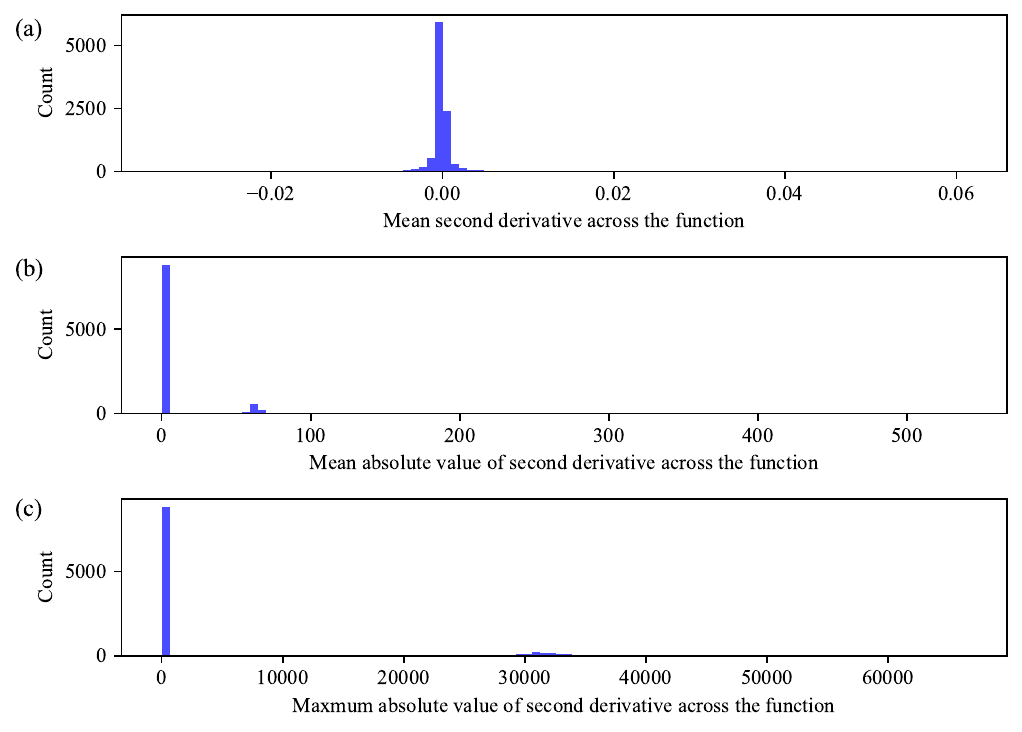}
    \caption{Distribution of second-order derivatives of functions $\fsij$. Includes all functions, regardless of whether they are linear, JumpReLU, or neither. For a version that only includes non-linear, non-JumpReLU functions, see Figure \ref{fig:second_derivative_distribution_nonlinear}.
        (a) The mean of the second-order derivative over the region of the input space.
        (b) The mean of the absolute value of the second-order derivative over the region of the input space.
        (c) The maximum value the second-order derivative takes in the region of the input space.
        Note that we are approximating the second derivative by looking at changes over a very small region (specifically $0.005$), i.e., we do not take the limit as the size of this small region goes to zero; this is important because derivatives which would otherwise be undefined or infinite become finite with this approximation and therefore can be shown on the histograms.
        Also, we note that the means and maxima are taken over the region of the input space in which SAE features exist; see the footnote on page \pageref{footnote:fsij_domain}.
    }
    \label{fig:second_derivative_distribution}
\end{figure}

\begin{figure}
    \centering
    \includegraphics{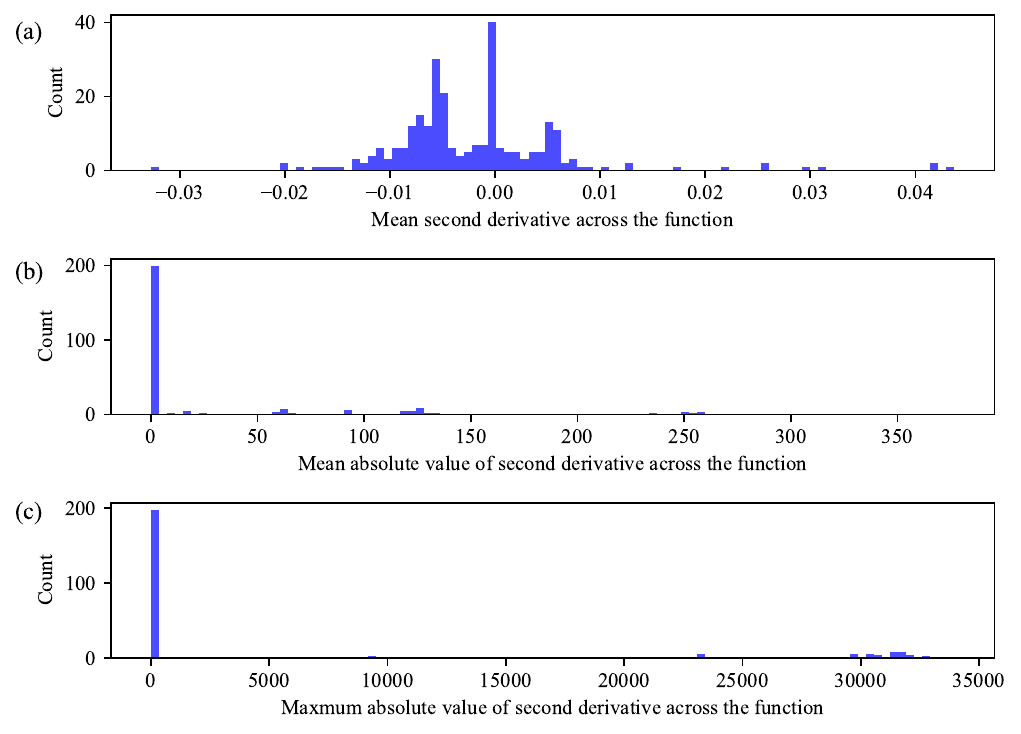}
    \caption{Distribution of second-order derivatives of functions $\fsij$. Unlike Figure \ref{fig:second_derivative_distribution}, this figure only includes the subset of the functions that are neither linear nor JumpReLU=like.
        (a) The mean of the second-order derivative over the region of the input space.
        (b) The mean of the absolute value of the second-order derivative over the region of the input space.
        (c) The maximum value the second-order derivative takes in the region of the input space.
        Note that we are approximating the second derivative by looking at changes over a very small region (specifically $0.005$), i.e. we do not take the limit as the size of this small region goes to zero; this is important because derivatives which would otherwise be undefined or infinite become finite with this approximation and therefore can be shown on the histograms.
        Also, we note that the means and maxima are taken over the region of the input space in which SAE features exist; see the footnote on page \pageref{footnote:fsij_domain}.
    }
    \label{fig:second_derivative_distribution_nonlinear}
\end{figure}

We can measure this more precisely by looking at the second-order derivative of $\fsij$.
A zero second-order derivative across the whole domain would imply a linear function and, therefore, perfect predictive power of the Jacobian, while the larger the absolute value of the second-order derivative, the less predictive the Jacobian will be.
This distribution is shown in Figure \ref{fig:second_derivative_distribution}.
The same distribution, which only includes the non-linear, non-JumpReLU functions, is shown in Figure \ref{fig:second_derivative_distribution_nonlinear}.
On average, the second derivative is extremely small for all features and effectively zero for the vast majority.

\begin{figure}
    \centering
    \includegraphics{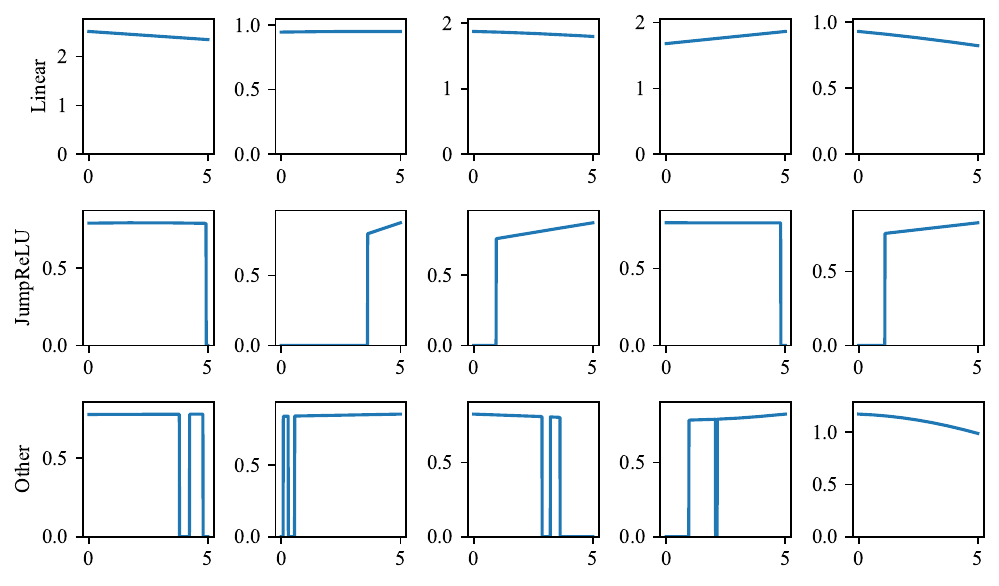}
    \caption{Additional examples of scalar functions between $\sxj$ to $\syi$. The top row shows linear functions, the middle row shows JumpReLU functions, and the bottom row shows other functions. Recall that linear functions constitute a majority of the functions we observe empirically and that using JSAEs instead of traditional SAEs further increases the proportion of linear functions.}
    \label{fig:func_examples}
\end{figure}

\begin{figure}
    \centering
    \includegraphics{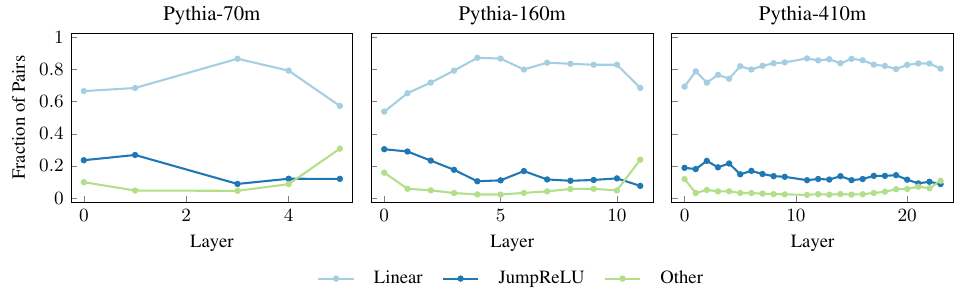}
    \caption{
        The fractions of Jacobian elements that exhibit a linear relationship between the input and output SAE latent activations, a JumpReLU-like relationship, and an uncategorized relationship, as described in Section~\ref{sec:mostly_linear}.
        Here, we consider Jacobian SAEs trained on the feed-forward network at different layers of Pythia-70m, 160m, and 410m with fixed expansion factors $R=64$ and $k=32$.
        We computed the fractions over 1 million samples.
    }
    \label{fig:linear_layer}
\end{figure}

\begin{figure}
    \centering
    \includegraphics{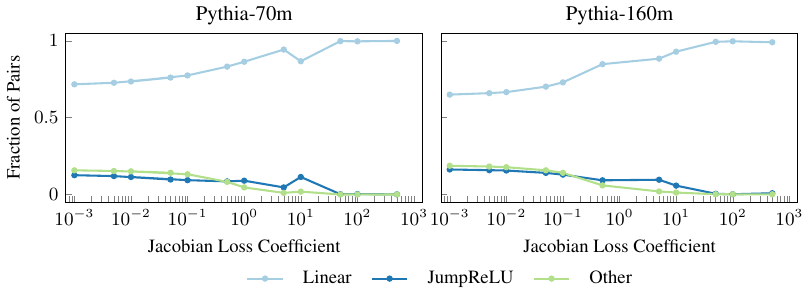}
    \caption{
        The fractions of Jacobian elements that exhibit a linear relationship between the input and output SAE latent activations, a JumpReLU-like relationship, and an uncategorized relationship, as described in Section~\ref{sec:mostly_linear}.
        Here, we consider Jacobian SAEs trained on the feed-forward network at layer 3 of Pythia-70m (left) and layer 7 of Pythia-160m (right), with fixed expansion factors $R=64$ and $k=32$ and varying Jacobian loss coefficient (Section~\ref{sec:methods}).
        We computed the fractions over 1 million samples.
    }
    \label{fig:linear_jacobian_coef}
\end{figure}

\begin{figure}
    \centering
    \includegraphics{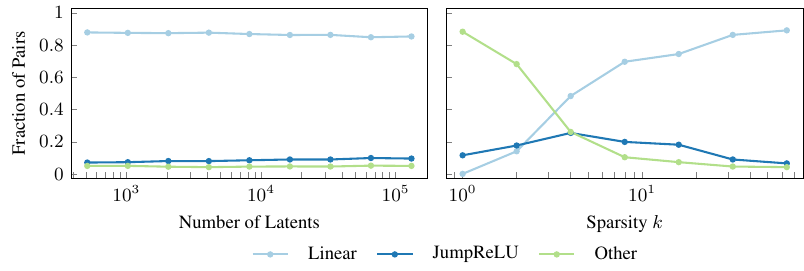}
    \caption{
        The fractions of Jacobian elements that exhibit a linear relationship between the input and output SAE latent activations, a JumpReLU-like relationship, and an uncategorized relationship, as described in Section~\ref{sec:mostly_linear}.
        Here, we consider Jacobian SAEs trained on the feed-forward network at layer 3 of Pythia-70m with varying expansion factors (and hence numbers of latents; left) but fixed sparsities $k=32$, and varying sparsities but fixed expansion factors $R=64$ (Section~\ref{sec:methods}).
        We computed the fractions over 1 million samples.
    }
    \label{fig:linear_expansion_factor_k}
\end{figure}

\section{Qualitative examples of the computations discovered by JSAEs}\label{app:qualitative}

A common approach to interpreting LLM components like neurons or SAE latents is to collect token sequences and the corresponding activations over a text dataset \citep[e.g.,][]{yun_transformer_2021,bills_language_2023}.
For example, the greatest latent activations may be retained, or activations from different quantiles of the distribution over the dataset \citep{bricken_monosemanticity_2023,choi_scaling_2024,paulo_automatically_2024}.

We determined the set of `top' output SAE latent indices by collecting the mean absolute values of non-zero Jacobian elements over a text dataset and sorting the output latents in descending order.
Then, for each output latent, we found the input SAE latents that were most strongly connected to the output latent, again by sorting the input latents in descending order of the mean absolute value of non-zero Jacobian elements over the dataset.
Finally, for both the output and input latents, we collected the individual latent activations over text samples with a context length of $16$ tokens, retaining samples where at least one token produced a non-zero activation for the SAE latent.
We chose a short context length to conveniently display the examples in a table format, and display here the top eight examples for each latent index, sorting the examples in descending order of the maximum latent activation over its tokens.

Each of the following figures comprises a table for a single output SAE latent (in pink), and a series of tables for the input latents with the greatest influences on the output latent, as determined by the mean absolute value of non-zero Jacobian elements.
Conceptually, one may consider each figure as describing a single `function', where the output and input latents represent the function output and inputs, respectively.
Each table within the figure of examples displays a list of at most $12$ examples, each comprising $16$ tokens; we exclude the end-of-sentence token for brevity.
The values of non-zero Jacobian elements and the activations of the corresponding input and output SAE latent indices are indicated by the opacity of the background color for each token.
We take the opacity to be the element or activation divided by the maximum value over the dataset, i.e., all the examples with a non-zero Jacobian element for a given pair of input and output SAE latent indices.
For clarity, we report the maximum element or activation alongside the colored tokens.

\renewcommand{\arraystretch}{0.5}
\renewcommand{\fboxsep}{0.5pt}

\input{feature_pairs_tex/out-pythia-410m-layer-15}

\section{Training}\label{app:training}

Our training implementation is based on the open-source SAELens library \citep{bloom_jbloomaus_2023}.
We train each pair of SAEs on 300 million tokens from the Pile \citep{gao_pile_2020}, excluding the copyrighted Books3 dataset, for a single epoch.
Except where noted, we use a batch size of 4096 sequences, each with a context size of 2048.
At a given time, we maintain 32 such batches of activation vectors in a buffer that is shuffled before training, which reduces variance in the training signal.

We use the Adam optimizer \citep{kingma_adam_2017} with the default beta parameters and a constant learning-rate schedule with 1\% warm-up steps, 20\% decay steps, and a maximum value of \qty{5e-4}.
Additionally, we use 5\% warm-up steps for the coefficient of the Jacobian term in the training loss.
We initialize the decoder weight matrix to the transpose of the encoder, and we scale the decoder weight vectors to unit norm at initialization and after each training step \citep{gao_scaling_2024}.

Except where noted, we choose an expansion factor $R=32$, keep the $k=32$ largest latents in the TopK activation function of each of the input and output SAEs, and choose a coefficient of $\lambda=1$ for the Jacobian term in the training loss.

\subsection{Training signal stability}
We initially considered the following setup:
\begin{equation}
  \sx = \encx(\x)\ ,\quad
  \hatx = \decx(\sx)\ ,\quad
  \y = f(\hatx)\ ,\quad
  \sy = \ency(\y)\ ,\quad
  \haty = \decy(\sy)
\end{equation}
The problem with this arrangement is that the second SAE depends on an output from the first SAE.
Since both SAEs are trained simultaneously, we found that this compromised training signal stability -- whenever the first SAE changed, the training distribution of the second SAE changed with it.
Additionally, at the start of training, when the first SAE was not yet capable of outputting anything meaningful, the second SAE had no meaningful training data at all, which not only made it impossible for the second SAE to learn but also made the first SAE less stable via the Jacobian sparsity loss term.

To address this problem, we instead used this setup:
\begin{equation}
    \sx = \encx(\x)\ ,\quad
    \hatx = \decx(\sx)\ ,\quad
    \y = f(\x)\ ,\quad
    \sy = \ency(\x)\ ,\quad
    \haty = \decy(\sy)
\end{equation}
Importantly, we pass the actual pre-MLP activations $\x$ rather than the reconstructed activations $\hatx$ into the MLP $f$.
In addition to improving training stability, we believe this setup to be more faithful to the underlying model because both SAEs are trained on the unmodified activations that pass through the MLP.




\section{Evaluation}
\label{app:evaluation}

We evaluated each of the input and output SAEs during training on ten batches of eight sequences, where each sequence has a context size of 2048, i.e., approximately 160K tokens.
We computed the sparsity of the Jacobian, measured by the mean number of absolute values above $0.01$ for a single token, separately after training.
In this case, we collected statistics over 10 million tokens from the validation subset of the C4 text dataset.

For reconstruction quality, we report the mean cosine similarity between input activation vectors and their autoencoder reconstructions, the explained variance (MSE reconstruction error divided by the variance of the input activation vectors), and the MSE reconstruction error.

For model performance preservation, we report the cross-entropy loss score, which is the increase in the cross-entropy loss when the input activations are replaced by their autoencoder reconstruction divided by the increase in the loss when the input activations are ablated (set to zero).

For sparsity, we report the number of `dead' latents that have not been activated (i.e., appeared in the $k$ largest latents of the TopK activation function) within the preceding 10 million tokens during training and the number of latents that have activated fewer than once per 1 million tokens during training on average.

Given an expansion factor of $64$, $k=32$, and a Jacobian loss coefficient of $1$, i.e., fixed hyperparameters, we find that the reconstruction error and cross-entropy loss score are consistently better for the input SAE than the output SAE.
Additionally, we find that the performance is generally poorer for the intermediate layers than early and later layers.

\begin{figure*}
    \centering
    \includegraphics{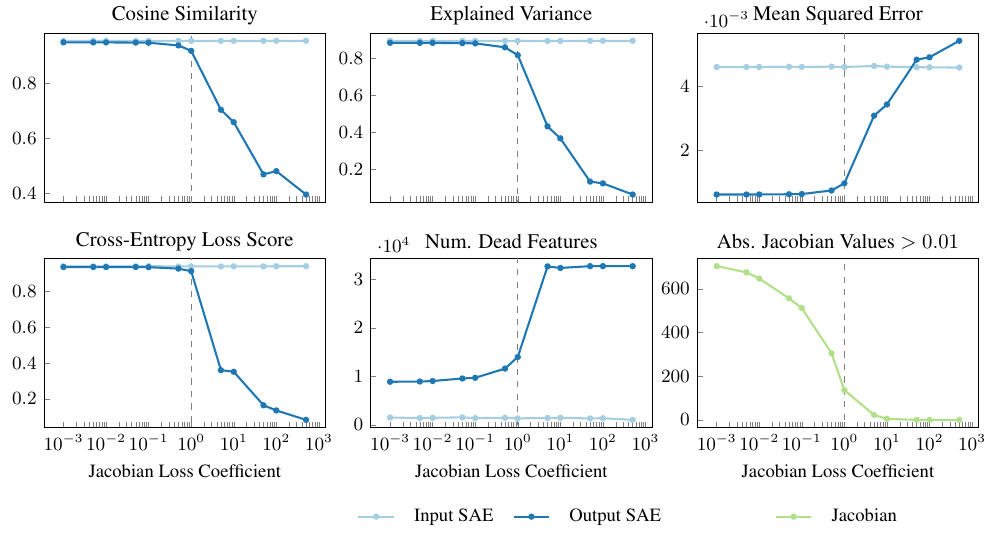}
    \caption{
        Reconstruction quality, model performance preservation, and sparsity metrics against the Jacobian loss coefficient.
        Here, we consider Jacobian SAEs trained on the feed-forward network at layer 3 of Pythia-70m with expansion factor $R=64$ and $k=32$.
        Recall that the maximum number of non-zero Jacobian values is $k^2=1024$.
        In accordance with Figure~\ref{fig:tradeoff_reconstr_sparsity_410m}, all evaluation metrics degrade for values of the coefficient above 1. See Appendix~\ref{app:evaluation} for details of the evaluation metrics.
    }
    \label{fig:jacobian_coef_70m}
\end{figure*}

We speculate that it is necessary to tune our hyperparameters for each layer individually to achieve improved performance; see, for example, Figures~\ref{fig:jacobian_coef_70m} and \ref{fig:jacobian_coef_160m} for the variation of our evaluation metrics against the coefficient of the Jacobian loss term for individual layers of Pythia-70m and 160m.

\begin{figure}
    \centering
    \includegraphics{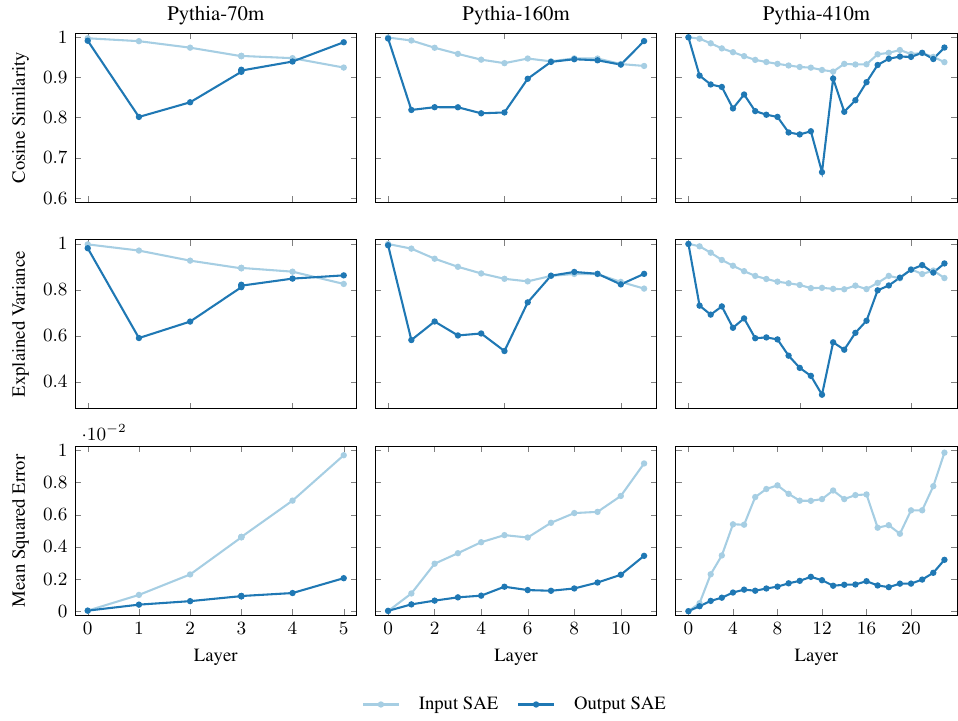}
    \caption{Reconstruction quality metrics for Jacobian SAEs trained on the feed-forward networks at every layer (residual block) of Pythia transformers. The cosine similarity is taken between the input and reconstructed activation vectors, and the explained variance is the MSE reconstruction error divided by the variance of the input activations. For each SAE, the expansion factor is $R=64$ and $k=32$; the Jacobian loss coefficient is 1.}
    \label{fig:reconstruction_quality}
\end{figure}

\begin{figure}
    \centering
    \includegraphics{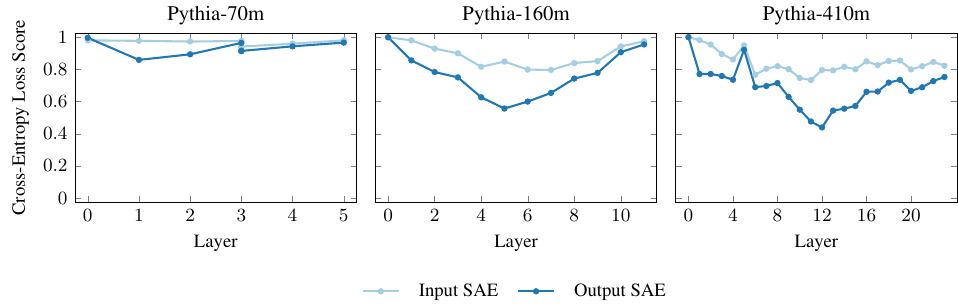}
    \caption{Model performance preservation metrics for Jacobian SAEs trained on the feed-forward networks at every layer (residual block) of Pythia transformers. The cross-entropy loss score is the increase in the cross-entropy loss when the input activations are replaced by their autoencoder reconstruction divided by the increase when the input activations are ablated (set to zero). For each SAE, the expansion factor is $R=64$ and $k=32$; the Jacobian loss coefficient is 1.}
    \label{fig:ce_loss_score}
\end{figure}

\begin{figure}
    \centering
    \includegraphics{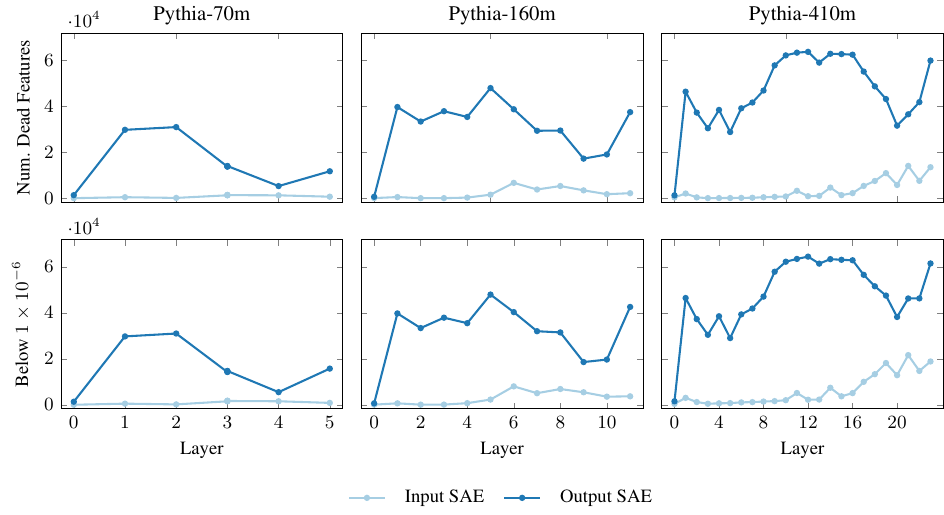}
    \caption{Sparsity metrics per layer for Jacobian SAEs trained on the feed-forward networks at every layer (residual block) of Pythia transformers. Recall that the $L^0$ norm per token for each of the input and output SAEs is fixed at $k$ by the TopK activation function. For each SAE, the expansion factor is $R=64$ and $k=32$; the Jacobian loss coefficient is 1.}
    \label{fig:sparsity}
\end{figure}

\begin{figure}
    \centering
    \includegraphics{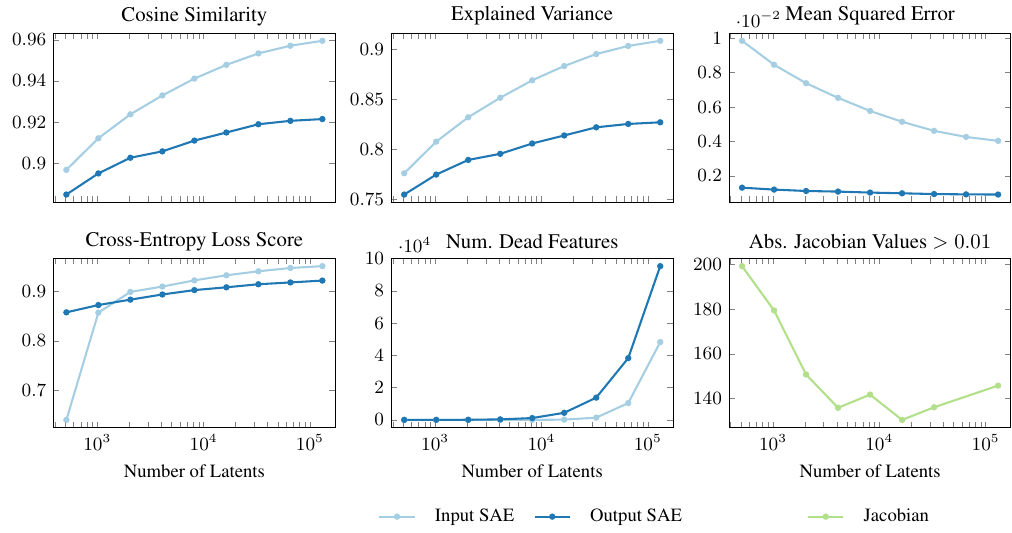}
    \caption{
        Reconstruction quality, model performance preservation, and sparsity metrics against the number of latents.
        Here, we consider Jacobian SAEs trained on the feed-forward network at layer 3 of Pythia-70m (model dimension 512) with $k=32$.
        Recall that the maximum number of non-zero Jacobian values is $k^2=1024$.
        The reconstruction quality and cross-entropy loss score improve as the number of latents increases, and the number of dead features grows more quickly for the output SAE than the input SAE.
        See Appendix~\ref{app:evaluation} for details of the evaluation metrics.
    }
    \label{fig:expansion_factor}
\end{figure}

\begin{figure}
    \centering
    \includegraphics{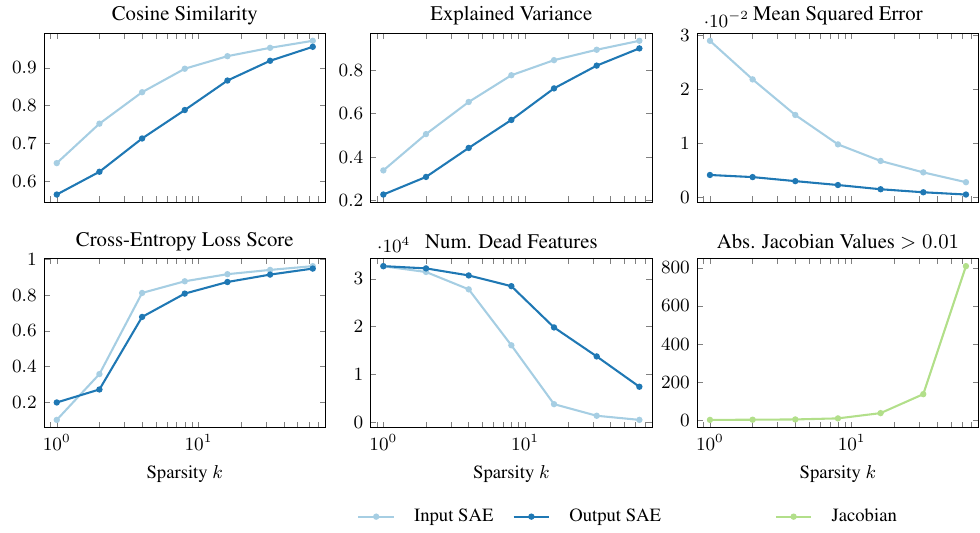}
    \caption{
        Reconstruction quality, model performance preservation, and sparsity metrics against the $k$ largest latents to keep in the TopK activation function.
        Here, we consider Jacobian SAEs trained on the feed-forward network at layer 3 of Pythia-70m with expansion factor $R=64$.
        Recall that the maximum number of non-zero Jacobian values is $k^2$.
        The reconstruction quality and cross-entropy loss score improve as $k$ increases, and the number of dead features decreases.
        See Appendix~\ref{app:evaluation} for details of the evaluation metrics.
    }
    \label{fig:k}
\end{figure}

\begin{figure*}
    \centering
    \includegraphics{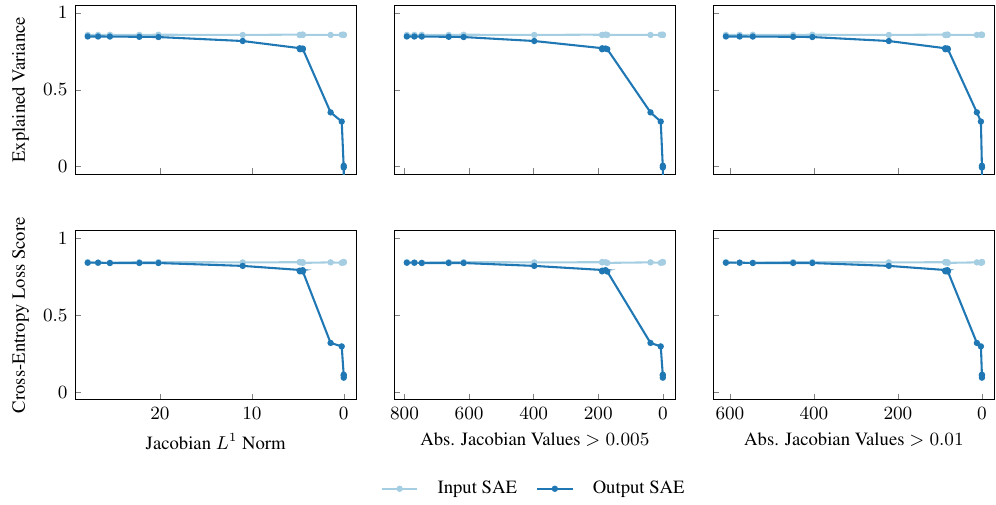}
    \caption{
        Pareto frontiers of the explained variance and cross-entropy loss score against different sparsity measures when varying the Jacobian loss coefficient.
        Here, we consider Jacobian SAEs trained on the feed-forward network at layer 3 of Pythia-70m with expansion factor $R=64$ and $k=32$.
        Recall that the maximum number of (dead) latents is $32768$ ($64$ times the model dimension $512$), and the maximum number of non-zero Jacobian values is $k^2=1024$.
        See Appendix~\ref{app:evaluation} for details of the evaluation metrics.
    }
    \label{fig:pareto_70m}
\end{figure*}

\begin{figure*}
    \centering
    \includegraphics{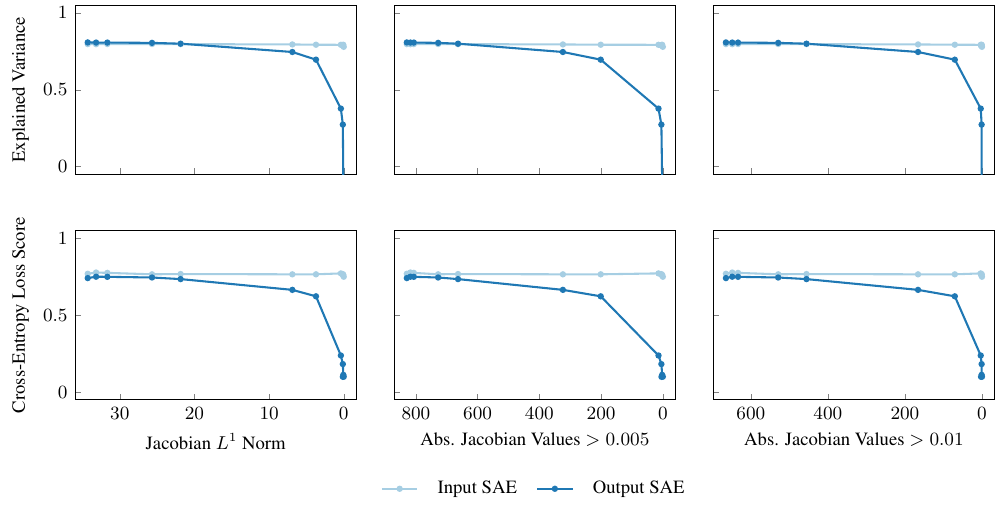}
    \caption{
        Pareto frontiers of the explained variance and cross-entropy loss score against different sparsity measures when varying the Jacobian loss coefficient.
        The coefficient has a relatively small impact on the reconstruction quality and sparsity of the input SAE, whereas it has a large effect on the sparsity of the output SAE and elements of the Jacobian matrix.
        Here, we consider Jacobian SAEs trained on the feed-forward network at layer 7 of Pythia-160m with expansion factor $R=64$ and $k=32$.
        Recall that the maximum number of (dead) latents is $49152$ ($64$ times the model dimension $768$), and the maximum number of non-zero Jacobian values is $k^2=1024$.
        See Appendix~\ref{app:evaluation} for details of the evaluation metrics.
    }
    \label{fig:pareto_160m}
\end{figure*}

\begin{figure*}
    \centering
    \includegraphics{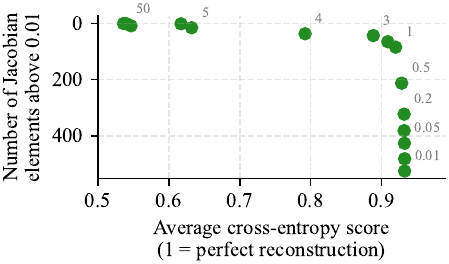}
    \caption{The trade-off between reconstruction quality and Jacobian sparsity as we vary the Jacobian loss coefficient. Each dot represents a pair of JSAEs trained with a specific Jacobian coefficient.
    Measured on layer 3 of Pythia-70m with $k=32$.}
    \label{fig:tradeoff_reconstr_sparsity_70m}
\end{figure*}

\section{More data on Jacobian sparsity}\label{app:jac_sparsity}
\begin{figure*}
    \centering
    \includegraphics[width=\linewidth]{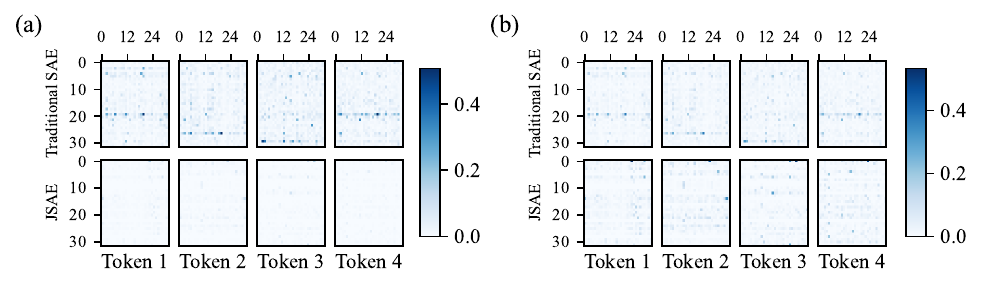}
    \caption{
    Comparison of Jacobians from traditional SAEs vs JSAEs, same as Figure \ref{fig:jac_sparsity} but with different normalization.
    (a) Not normalized.
    (b) L2 normalized.
    Measured on layer 15 of Pythia-410m.}
    \label{fig:jac_examples_normed}
\end{figure*}

\begin{figure*}
    \centering
    \includegraphics[width=\linewidth]{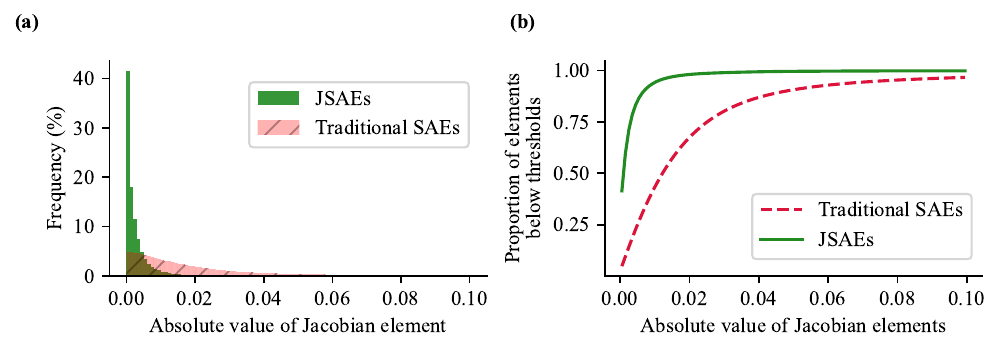}
    \caption{
    Further data showing that JSAEs induce much greater Jacobian sparsity than traditional SAEs.
    (a) A histogram of the absolute values of Jacobian elements in JSAEs versus traditional SAEs.
    JSAEs induce significantly more sparse Jacobians than standard SAEs.
    This means that there is a relatively small number of input-output feature pairs which explain a very large fraction of the computation being performed.
    Note that only the $k\times k$ elements corresponding to active latents are included in the histogram -- the remaining $(\dimsy-k)\times(\dimsx-k)$ elements are zero by definition both for JSAEs and standard TopK SAEs.
    The histogram was collected over 10 million tokens from the validation subset of the C4 text dataset, which produced 10.24 billion feature pairs.
    (b) The cumulative distribution function of the absolute values of Jacobian elements, again demonstrating that JSAEs induce significantly more computational sparsity than traditional SAEs.
    Measured on layer 15 of Pythia-410m.}
    \label{fig:jac_hist}
\end{figure*}

In Figure \ref{fig:sparsity} we showed that Jacobians are much more sparse with JSAEs than traditional SAEs.
To this end, we provided a representative example of what the Jacobians look like with JSAEs vs traditional SAEs.
Some readers may object that this is not an apples-to-apples comparison since JSAEs are optimizing for lower L1 on the Jacobian, so it may be the case that JSAEs merely induce Jacobians with smaller elements, but their distribution may still be the same.
To address this criticism, the examples are L2 normalized; we provide un-normalized versions as well as L1 normalized versions of the example Jacobians in Figure \ref{fig:jac_examples_normed}.
We also provide a histogram and a CDF of the distribution of absolute values of Jacobian elements in Figure \ref{fig:jac_hist}, which is taken across 10 million tokens.

\subsection{Jacobian norms}\label{app:jac_norms}
\begin{figure*}
    \centering
    \includegraphics[width=\linewidth]{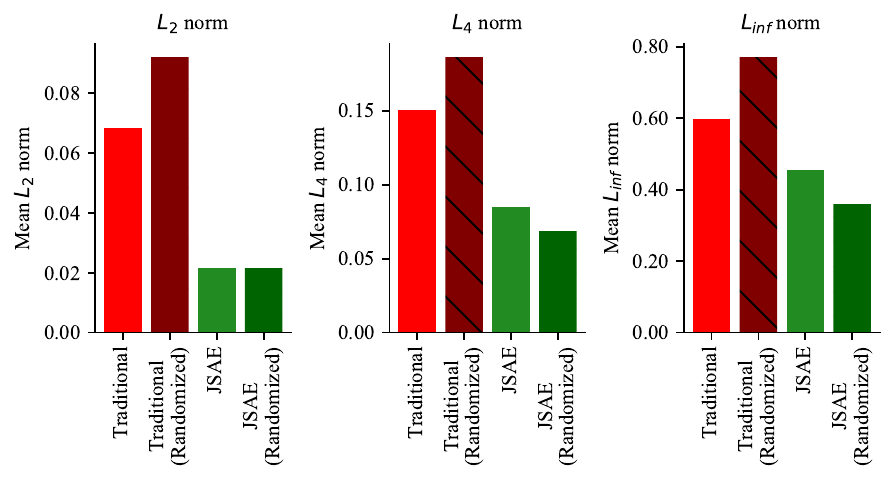}
    \caption{$L_p$ norms of the Jacobians. We measure these by flattening the Jacobians and treating them as a vector. These results imply that the Jacobians are in fact becoming more sparse, as opposed to merely becoming smaller (see Section~\ref{app:jac_norms}). Averaged across 1 million tokens, measured on layer 3 of Pythia-70m.}
    \label{fig:jac_norms}
\end{figure*}

In this section, we address an objection we expect some readers will have to our measures of sparsity.
Our main metric for sparsity is the percentage of elements with absolute values above certain small thresholds (e.g. Figure~\ref{fig:jac_sparsity}).
However, one can imagine two distributions with the same degree of sparsity, but vastly different results on this metric due to a different standard deviation.
For instance, imagine two Gaussian distributions, both with $\mu=0$ but with significantly different standard deviations, $\sigma_1\gg\sigma_2$.
They would score very differently on our metric, but their degrees of sparsity would not be meaningfully different (since sparsity requires there to be a small handful of relatively large elements).
Since our $L_1$ penalty encourages the Jacobians to be smaller, it could be that they simply become more tightly clustered around 0.
However, this is not the case.
We can measure this by looking at the "norms" of the Jacobian, i.e. we flatten the Jacobian, treat it as a vector, and compute its $L_p$ norms.
If the Jacobian is merely becoming smaller, we would expect all of its $L_p$ norms to decrease at roughly the same rate.
On the other hand, if the Jacobian is becoming sparser, we would expect its $L_1,L_2$ norms to decrease while its $L_4,\dots,L_\infty$ norms, which depend more strongly on the presence or absence of a few large elements, should stay roughly the same.
We present these results in Figure~\ref{fig:jac_norms}, as we can see, the Jacobian does become slightly smaller, but most of the effect we see is indeed the Jacobian becoming significantly more sparse.

\begin{figure*}
    \centering
    \includegraphics[width=\linewidth]{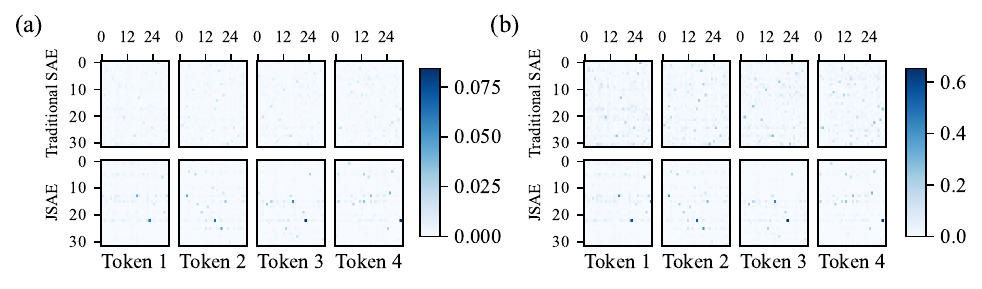}
    \caption{
    Comparison of Jacobians from traditional SAEs vs JSAEs, same as Figure \ref{fig:jac_sparsity} but with different normalization.
    (a) L1 normalized.
    (b) L2 normalized.
    Measured on layer 3 of Pythia-70m.}
    \label{fig:jac_examples_normed_70m}
\end{figure*}

\begin{figure*}[t]
    \centering
    \includegraphics{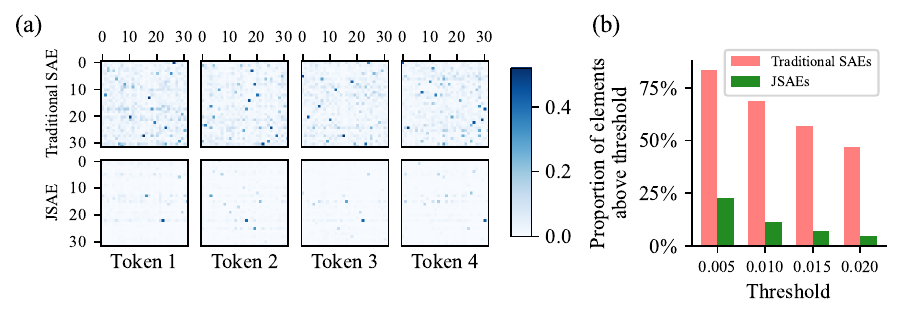}
    \caption{
    JSAEs induce a much greater degree of sparsity in the elements of the Jacobian than traditional SAEs.
    Identical to Figure~\ref{fig:jac_sparsity} but measured on layer 3 of Pythia-70m.
    }
    \label{fig:jac_sparsity_70m}
\end{figure*}

\begin{figure}
    \centering
    \begin{subfigure}{0.49\textwidth}
        \centering
        \includegraphics{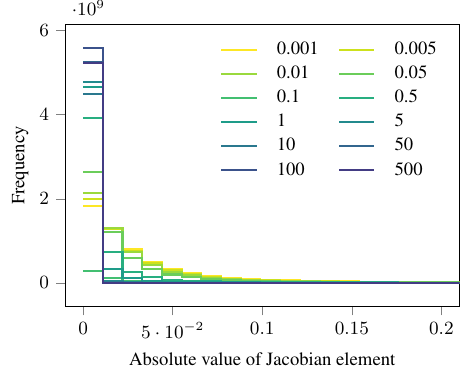}
        \caption{Pythia-70m Layer 3}
    \end{subfigure}
    \begin{subfigure}{0.49\textwidth}
        \centering
        \includegraphics{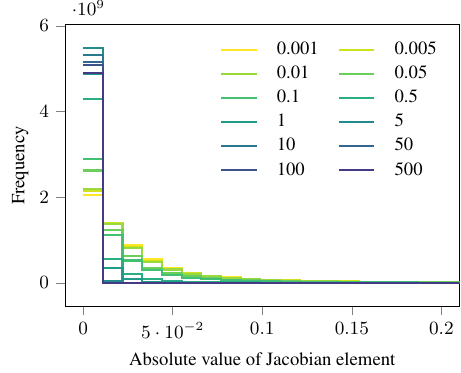}
        \caption{Pythia-160m Layer 7}
    \end{subfigure}
    \caption{
        Histograms that show the frequency of absolute values of non-zero Jacobian elements for different values of the coefficient of the Jacobian loss term.
        As the coefficient increases, the frequency of larger values decreases, i.e., the Jacobian becomes sparser.
        We provide further details in Figure~\ref{fig:jac_hist}.
    }
    \label{fig:jac_hist_multiple_models}
\end{figure}

\begin{figure*}
    \centering
    \includegraphics[width=0.5\linewidth]{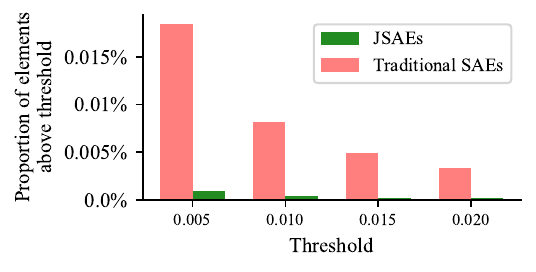}
    \caption{The Jacobians aren't only sparse locally (i.e. on each token in each prompt), but also globally (i.e. when averaged across many tokens), much more so than with traditional SAEs.
    In particular, here we consider the full $\dimsy\times\dimsx$ Jacobian (i.e. not slicing based on the TopK), which we average across 10 million tokens ($\frac{1}{N}\sum_\text{prompt,token}\mat{J}$) before considering its summary statistics.
    This is an important measure as it confirms that the connections found by JSAEs are indeed sparse in a global sense, not just when conditioning on a specific model input.
    Measured on layer 15 of Pythia-410m.
    Note that the small numbers on the y-axis are due to the fact that, unlike in e.g. Figure~\ref{fig:jac_sparsity}, here we set 100\% to be $\dimsy\times\dimsx$ rather than $k\times k$.
    We also note that for each element in the Jacobian, we are only taking the average over the tokens on which the corresponding output SAE latent is selected by the TopK activation function (i.e. when at least one element in the row of the Jacobian is nonzero); this is important because otherwise this measure would significantly conflate the sparsity of the Jacobian itself with the sparsity of the activations of each individual latent.}
    \label{fig:jac_sparsity_global}
\end{figure*}

\begin{figure}
    \centering
    \includegraphics{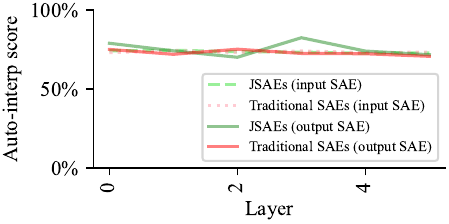}
    \caption{Automatic interpretability scores of JSAEs are very similar to traditional SAEs. Measured on all layers of Pythia-70m using the ``fuzzing'' scorer from \citet{paulo_automatically_2024}.}
    \label{fig:autointerp_70m}
\end{figure}

\begin{figure}
    \centering
    \includegraphics{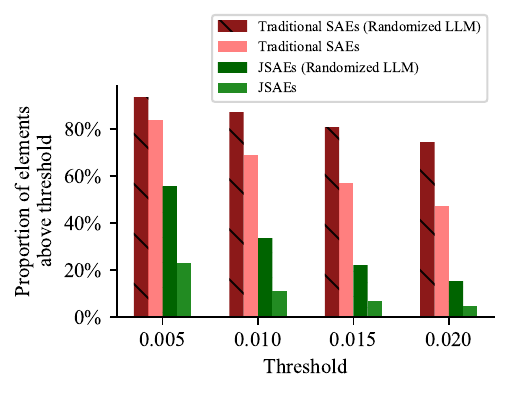}
    \caption{
    Jacobians are substantially more sparse in pre-trained LLMs than in randomly initialized transformers.
    This holds both when you actively optimize for Jacobian sparsity with JSAEs, and when you don't optimize for it and use traditional SAEs.
    The proportion of Jacobian elements with absolute values above certain thresholds.
    The figure shows the proportion of Jacobian elements with absolute values above certain thresholds.
    Identical to Figure \ref{fig:randomized_410m} but measured on layer~3 of Pythia-70m.}
    \label{fig:randomized_70m}
\end{figure}

\begin{figure*}
    \centering
    \includegraphics{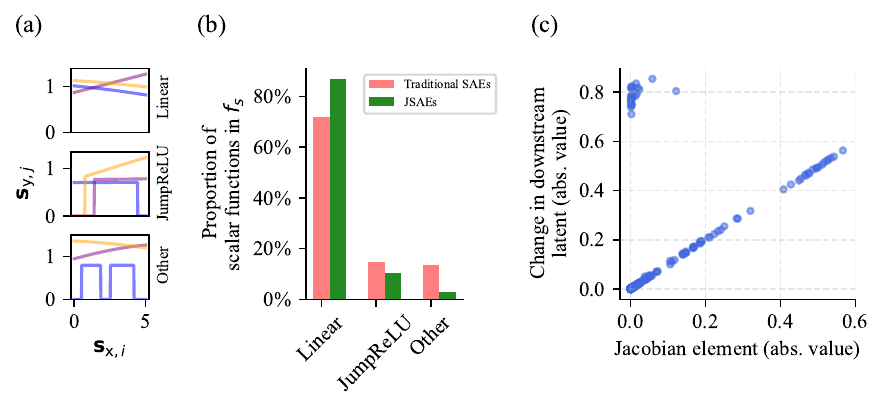}
    \caption{The function $f_s$, which combines the decoder of the first SAE, the MLP, and the encoder of the second SAE, is mostly linear.
    Identical to Figure~\ref{fig:mostly_linear_410m} but measured on layer 3 of Pythia-70m.
    }
    \label{fig:mostly_linear_70m}
\end{figure*}

\end{document}